\documentclass[sigconf]{aamas}

\usepackage{preamble}

\newif\ifarxiv
\arxivtrue 

\setcopyright{ifaamas}
\acmConference[AAMAS '24]{Proc.\@ of the 23rd International Conference
on Autonomous Agents and Multiagent Systems (AAMAS 2024)}{May 6 -- 10, 2024}
{Auckland, New Zealand}{N.~Alechina, V.~Dignum, M.~Dastani, J.S.~Sichman (eds.)}
\copyrightyear{2024}
\acmYear{2024}
\acmDOI{}
\acmPrice{}
\acmISBN{}

\acmSubmissionID{347}

\title{\iiql: Imitation Learning of Intent-Driven Expert Behavior}

\author{Sangwon Seo}
\affiliation{
  \institution{Rice University}
  \city{Houston}
  \country{USA}}
\email{sangwon.seo@rice.edu}

\author{Vaibhav Unhelkar}
\affiliation{
  \institution{Rice University}
  \city{Houston}
  \country{USA}}
\email{vaibhav.unhelkar@rice.edu}



\makeatletter
\gdef\@copyrightpermission{
	\begin{minipage}{0.3\columnwidth}
		\href{https://creativecommons.org/licenses/by/4.0/}{\includegraphics[width=0.90\textwidth]{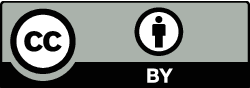}}
	\end{minipage}\hfill
	\begin{minipage}{0.7\columnwidth}
		\href{https://creativecommons.org/licenses/by/4.0/}{This work is licensed under a Creative Commons Attribution International 4.0 License.}
	\end{minipage}
	\vspace{5pt}
}
\makeatother

\begin{abstract}
When faced with accomplishing a task, human experts exhibit intentional behavior.
Their unique intents shape their plans and decisions, resulting in experts demonstrating diverse behaviors to accomplish the same task.
Due to the uncertainties encountered in the real world and their bounded rationality, experts sometimes adjust their intents, which in turn influences their behaviors during task execution.
This paper introduces \iiql, a novel imitation learning algorithm to mimic these diverse intent-driven behaviors of experts.
Iteratively, our approach estimates expert intent from heterogeneous demonstrations and then uses it to learn an intent-aware model of their behavior.
Unlike contemporary approaches, \iiql is capable of addressing sequential tasks with high-dimensional state representations, while sidestepping the complexities and drawbacks associated with adversarial training (a mainstay of related techniques).
Our empirical results suggest that the models generated by \iiql either match or surpass those produced by recent imitation learning benchmarks in metrics of task performance.
Moreover, as it creates a generative model, \iiql demonstrates superior performance in intent inference metrics, crucial for human-agent interactions, and aptly captures a broad spectrum of expert behaviors.
\end{abstract}

\begin{document}
\pagestyle{fancy}
\fancyhead{}

\maketitle
\section{Introduction}
With the growing demand for AI systems that can seamlessly interact and emulate human behavior, modeling human behavior and learning thereof have been important problems in the AI community~\cite{albrecht2018autonomous,argall2009survey,ivanovic2018generative}.
For instance, when developing an autonomous agent, behavior of human experts serves as a useful starting point, especially for the tasks where humans excel.
Learning from human experts proves especially beneficial in scenarios where rewards are sparse or ill-defined, rendering policy learning through reinforcements impractical~\cite{ho2016generative,ziebart2008maximum}.
Modeling human behavior is also essential in human-agent interaction contexts, as agents (i.e. autonomous vehicles and robots) have to predict and reason about human behavior for fluent coordination \cite{hoffman2019evaluating,hiatt2017human,unhelkar2020semi,unhelkar2020decision}.
Without an accurate model of human behavior, robots  misinterpret human intentions and can cause sub-optimal human-robot interaction.
\ifarxiv
\blfootnote{This article is an extended version of an identically-titled paper accepted at the \textit{International Conference on Autonomous Agents and Multiagent Systems (AAMAS 2024)}.}
\fi

\begin{figure}[t]
\centering
\includegraphics[width=0.9\columnwidth]{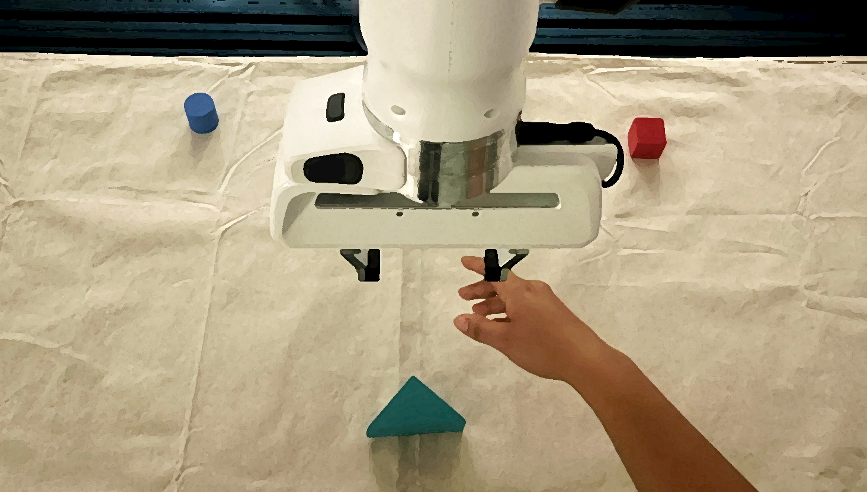}
\caption{Consider the task of emptying a table. Different individuals may accomplish this task differently; some starting with the red block, while others with the green or blue. Intent-driven imitation learning aims to model this diversity in behaviors (arising from differences in  experts' intents) from heterogeneous demonstrations.
}
\label{fig. teaser}
\end{figure}

Imitation learning, also known as learning from demonstrations, is a widely-used data-driven paradigm to model human behavior in sequential decision-making tasks~\cite{abbeel2004apprenticeship,ho2016generative,ziebart2008maximum,arora2021survey}.
Algorithms for imitation learning (IL) most commonly utilize \textit{observation-action} trajectories of human experts' task executions and return an estimate of true expert behavior (denoted as $\pi_E$).
Mathematically, the behavior is learned in the form of a conditional distribution $\pi_E \approx \pi(a|s)$, where the variable $a$ denotes task-specific actions and $s$ denotes the \textit{observable} task context.
In other words, the traditional IL paradigm assumes expert behavior as being influenced solely by observable and known decision-making factors.

However, as motivated in \autoref{fig. teaser}, the reality is more nuanced.
Different experts exhibit varying behaviors for the same task.
Furthermore, even a single expert's behavior can fluctuate markedly based on their \textit{intents} for identical task contexts.
In this sense, intent of a human expert encompasses her preferred approach to executing a task, particularly when it can be accomplished in a variety of ways.
Moreover, experts often modify their intent during tasks to cater to evolving scenarios. 
Hence, intent emerges as a \textit{latent decision-making factor} upon which broader, high-level task strategies are formed \cite{byrne1998learning}.
We posit that \textit{recognizing and factoring intent and its dynamic nature is crucial for accurate modeling of human behavior}.
Given that intent is a cognitive construct and inherently unobservable, conventional imitation learning approaches (which assume behavior depends only on the observable context) fall short in addressing this complexity.

Informed by this research gap, we consider the problem of learning \textbf{intent-driven expert behavior}.
Mathematically, we seek to estimate $\pi_E \approx \pi(a|s,x)$, where $x$ denotes the intent of the human expert, from a set of heterogeneous expert demonstrations.
In addition to utilizing diverse demonstrations without segmentation,
learning intentional behavior offers two distinct advantages over traditional imitation learning.
First, \textit{intent-driven models can be leveraged to discern the intents of human experts either during or post-task}.
Such capability is invaluable for assistive robots and autonomous agents, who can use these inferred intents to anticipate experts' actions over extended periods, enabling improved coordination \cite{lasota2017survey, thomaz2016computational, hiatt2017human}.
Secondly, \textit{intent-driven behavior models $\pi(a|s,x)$ can capture variety of expert behaviors and thus serve as a more comprehensive generative model of expert behavior} relative to intent-unaware models $\pi(a|s)$.
Representative generative models of human decision-making behavior are particularly useful in a variety of applications, such as development of digital twins.

Recognizing the diversity observed in expert behaviors, several imitation learning methods have been developed in recent years to learn a behavioral policy that depends on both the task context and latent decision-making factors~\cite{jing2021adversarial,sharma2018directed,unhelkar2019learning,orlov2022factorial,lee2020learning}.
However, as reviewed in Sec.~\ref{sec. related works} and demonstrated in Sec.~\ref{sec. experiments}, existing intent-aware methods either are designed for discrete domains and, thus, cannot scale up to tasks with large or continuous state spaces~\cite{unhelkar2019learning,orlov2022factorial}; or rely on generative adversarial training, whose performance is often unstable in practice~\cite{jing2021adversarial,lee2020learning,sharma2018directed}.

Towards addressing these limitations, we introduce \iiql: a novel algorithm for imitation learning of intent-driven expert behavior.
\iiql learns a generative model of heterogeneous expert behaviors, composed of an intent-aware estimate of the expert policy $\pi_E \approx \pi(a|s,x)$ and a model of experts' intent dynamics $\zeta(x'|s,x)$.
To realize stable learning in tasks with large or continuous state spaces, \iiql builds upon recent results from classical imitation learning~\cite{garg2021iq}.
Next, beginning with problem formulation, we describe \iiql and derive its convergence properties.
This analytical investigation is followed by a suite of experiments, which empirically compare \iiql to recent imitation learning baselines.
Our experiments confirm that \iiql is successfully able to estimate expert intent, learn predictive models of expert behavior, and generate diverse interpretable set of (both seen and unseen) expert behaviors.

\subsection{Motivating Scenario}
\label{sec. motivating scenario}
To further motivate the problem and describe our solution, we consider a 2-dimensional version of the task shown in \autoref{fig. teaser}.
In this task, which we call \twodim-$n$, where a human is tasked with visiting $n$ landmarks\footnote{Such tasks occur commonly in domains such as robotic manufacturing, as a component of more complex sorting, binning, and assembly tasks.}; however, the order to visit the landmarks is not specified.
As such, there are multiple ways to accomplish the task optimally and different individuals may accomplish this task differently.
Based on their intent, some experts may start with the red block, while others with the green or blue, and so on.
While expert behavior depends on both their starting location and intent, only the former is readily observable.
We seek algorithms that explicitly model and learn intent-driven policy of expert behavior from a set of heterogeneous expert demonstrations.
 
\subsection{Preliminaries}
To formulate the problem, we limit our scope to sequential tasks that can be described as Markov Decision Processes and expert behaviors that can be represented as an Agent Markov Model. 

\subsubsection{Markov Decision Processes (MDP)}
An MDP represents a sequential task via the tuple $\mathcal{M}\doteq(S, A, T, R, \mu_0, \gamma)$, where $S$ denotes the set of task states $s$, $A$ denotes the the set of actions, $T:S\times A \times S \rightarrow [0, 1]$ denotes the state transition probabilities, $\mu_0:S \rightarrow [0, 1]$ denotes the initial state distribution, $R: S\times A \rightarrow \mathbb{R}$ denotes the task reward, and $\gamma \in (0, 1)$ is the discount factor.
For our running example, the state denotes the agent location and action denotes the direction in which they intend to move next.
The solution to an MDP is denoted as policy, $\pi$, which prescribes the agent's next action. 
The degree of optimality of a policy is quantified through its expected cumulative reward $\bar{G}_\pi \doteq E_{\pi} [\sum_{t=0}^\infty \gamma^t R_t]$. 
An expert solving an MDP seeks to maximize $\bar{G}_\pi$ by solving for and executing the optimal policy $\pi^* \doteq \arg\max_\pi \bar{G}_\pi$.
Due to the Markovian property, there exists at least one optimal policy that is Markovian.
Mathematically, this Markovian optimal policy can be summarized via the conditional distribution $\pi^*(a|s):S\times A \rightarrow [0, 1]$, which denotes the probability of selecting an action $a$ in a given state $s$. 

\subsubsection{Imitation Learning.}
A Markovian policy $\pi(a|s)$ induces a stationary distribution of states and actions, referred to as the normalized occupancy measure 
\begin{equation}
\rho(s, a) \doteq (1-\gamma) \sum_{t=0}^\infty \gamma^t p(s^t = s, a^t = a|\pi, \mathcal{M}),    
\end{equation}
where $p(s^t, a^t|\pi, \mathcal{M})$ is the probability of being in state $s$ and taking action $a$ at time $t$ when following the policy $\pi$ in the task $\mathcal{M}$ \cite{puterman2014markov}. 
Conventionally, imitation learning seeks to learn a policy $\pi$ that matches the behavior of an expert from demonstrations $D$:
\begin{equation}
   \min_\pi L(\pi, \pi_E; D) 
\end{equation}
where $\pi_E$ is the expert's policy that is unknown to the imitation learner.
Due to the one-to-one correspondence between a policy and its occupancy measure, policy matching problem can be reformulated as an occupancy measure matching problem \cite{ho2016generative,ghasemipour2020divergence}:
\begin{align}
    \arg\min_\pi D_f \left(\rho_\pi (s, a) \middle|\middle| \rho_{\pi_E}(s, a)\right) \label{eq. occupancy measure matching}
\end{align}
where $D_f$ denotes $f$-Divergence, and $\rho_\pi$ and $\rho_{\pi_E}$ are the occupancy measures induced by the imitation learner and expert respectively.

\subsubsection{Agent Markov Model.}
While optimal behavior in MDP can be represented simply as a function of the state, in practice, expert behavior -- especially that of humans -- often hinges on additional decision-making factors.
As exemplified by our motivating scenario, several optimal policies might exist in the real world.
Human experts tend to show individual inclinations in choosing among these optimal policies.
The Agent Markov Model (AMM) provides a principled approach to represent these variety of behaviors exhibited by agents while solving sequential tasks~\cite{unhelkar2019learning}.
An AMM models behavior to depend both on the task state and latent states (e.g., trust, intent, and workload).
Mathematically, for a task described as an MDP $\mathcal{M}$, an Agent Markov Model (AMM) defines the behavior of agents as a tuple $(X, \zeta, b, \pi)$, where $X$ is the set of latent states, $\zeta:X \times S \times X \rightarrow [0, 1]$ is the transition probabilities of the latent states, $b:S \times X \rightarrow [0, 1]$ is the initial distribution of the latent states, and $\pi: S \times X \times A \rightarrow [0,1]$ is the agent policy.
Hence, actions of an AMM agent depend on both the task and latent states $a \sim \pi(a|s,x)$.
Further, the agent's latent state may change during the task execution based on the corresponding transition probability, $\zeta(x|s, x^-)$.

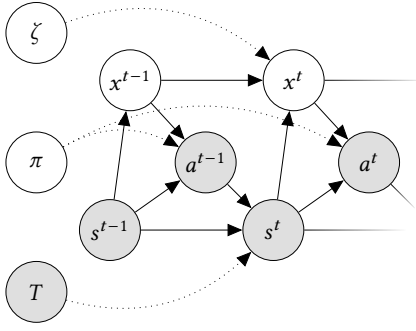
\begin{figure}[t]
    \centering
    \scalebox{0.9}{
    \begin{tikzpicture}
        \node[obs, minimum size=0.9cm] (st) {$s^t$} ;
        \node[obs,draw=none,fill=none, right=1.5cm of st, minimum size=0.9cm] (stn) {} ;
        \node[obs, left=1.5cm of st, minimum size=0.9cm] (stp) {$s^{t-1}$} ;        

        \node[obs, right=0.5cm of st, yshift=1cm, minimum size=0.9cm] (a) {$a^t$} ;
        \node[obs, right=0.5cm of stp, yshift=1cm, minimum size=0.9cm] (ap) {$a^{t-1}$} ;

        \node[latent, above=1.3cm of st, xshift=0.3cm, minimum size=0.9cm] (x) {$x^t$} ;
        \node[latent,draw=none, above=1.3cm of stn, xshift=0.3cm, minimum size=0.9cm] (xn) {} ;
        \node[latent, above=1.3cm of stp, xshift=0.3cm, minimum size=0.9cm] (xp) {$x^{t-1}$} ;

        \node[latent, left=4cm of a, minimum size=0.9cm] (pi) {$\pi$} ;
        \node[latent, above=1cm of pi, minimum size=0.9cm] (zeta) {$\zeta$} ;
        \node[obs, below=1cm of pi, minimum size=0.9cm] (tr) {$T$} ;

        \edge {stp} {ap} ;
        \edge {st} {a} ;
        \edge {ap} {st} ;
        \edge[-,path fading=east] {a} {stn} ;
        \edge {stp} {st} ;
        \edge[-,path fading=east] {st} {stn} ;
        \edge {xp} {ap} ;
        \edge {x} {a} ;
        \edge {xp} {x} ;
        \edge[-,path fading=east] {x} {xn} ;
        \edge {stp} {xp} ;
        \edge {st} {x} ;
        \path (pi) edge [->,dotted, bend left](a) ;
        \path (pi) edge [->,dotted, bend left](ap) ;
        \path (zeta) edge [->,dotted, bend left](x) ;
        \path (tr) edge [->,dotted, bend right](st) ;

    \end{tikzpicture}
    }
    \caption{Dynamic Bayesian network representing Intent-Driven Behavior. Shaded nodes denote known or observable variables; other variables are latent.} 
    \label{fig.graph_model}
\end{figure}

\subsection{Problem Formulation}
\label{sec. problem}
We now formalize the problem of learning intent-driven expert behavior.
Similar to the conventional imitation learning setting, we focus on sequential tasks that can be described as MDPs.
However, instead of assuming that expert behavior depends on the task state alone, we model it also depend on their time-varying intent.

\subsubsection{Model of Expert Behavior}
Mathematically, we model the expert as a special case of AMM in which the latent state $(x)$ corresponds to the expert intent. 
A dynamic Bayesian network for this representation is depicted in \autoref{fig.graph_model}.
Similar to \citet{seo2022semi}, we assume $X$ the set of intents is known and finite. 
Following \citet{jing2021adversarial}, we introduce the notation $\#$ to represent the imaginary value of the intent at time $t=-1$ for the sake of notational brevity.
By defining $X^+ \doteq X \cup \{\#\}$ and representing the initial intent distribution as $\zeta(x|s, x^-=\#) \doteq b(x|s)$, we redefine the intent transition probabilities to take the form $\zeta:X^+ \times S \times X \rightarrow [0, 1]$.
As $\zeta$ now encompasses the initial intent distribution and $X$ is known, we henceforth refer to the AMM model describing the expert as $\mathcal{N}_E$ as the reduced tuple $(\pi_E, \zeta_E)$.

\subsubsection{Inputs and Outputs}
Although intents influence expert behavior, they cannot be readily sensed or measured by imitation learners. 
Thus, we assume an expert demonstration does not include the information of intents $x$.
We define a demonstration as $\tau \doteq (s^{0:h}, a^{0:h})$ and a set of $d$ demonstrations as $D = \{\tau_m\}_{m=1}^d$, where $h$ is a task horizon.
In addition, we denote the label of intents for the $m$-th demonstration as $x_m^{0:h}$.
Formally, our problem assumes the following inputs: a task model $\mathcal{M}\setminus R$, a set of intents $X$, a set of expert demonstrations $D$, and optionally labels of intents for $l(\leq d)$ demonstrations $\{x_m^{0:h}\}_{m=1}^l$. 
Given these inputs, our goal is to learn an AMM model $\mathcal{N} = (\pi, \zeta)$ that mimics expert behavior.

\subsubsection{Objective}
Since the learner only has access to the observable part of the expert behavior $\{s, a\}$, we can consider simply matching the occupancy measure of the observable variables: 
\begin{align*}
    \arg\min_\pi D_f \left(\rho_\mathcal{N} (s, a) \middle|\middle| \rho_{{\mathcal{N}_E}}(s, a)\right) 
\end{align*}
However, under this criteria, the learned model could be far from the expert behavior, since multiple AMM models can exhibit the same $(s, a)$-occupancy measure~\cite{jing2021adversarial}.
To mitigate this ambiguity, hence, we define the normalized intent-aware occupancy measure: 
\begin{equation*}
\rho_{\mathcal{N}}(s, a, x, x^-) {\doteq} (1-\gamma) \sum_{t=0}^\infty \gamma^t p(s^t{=}s, a^t{=}a, x^t{=}x, x^{t-1}{=}x^-|\mathcal{N}, \mathcal{M})    
\end{equation*}
and pose the problem of learning intent-driven expert behavior as the following occupancy measure matching problem:
\begin{gather}
    \min_{\mathcal{N}} L_{saxx^-}(\mathcal{N}) {\doteq} \min_{\mathcal{N}} D_f \left(\rho_{\mathcal{N}}(s, a, x, x^-) \middle|\middle| \rho_{\mathcal{N}_E} (s, a, x, x^-)\right) \label{eq. main objective}.
\end{gather}

\section{Learning Intent-Driven Behavior}
\label{sec. method}
To effectively solve the optimization problem of Eq.~\ref{eq. main objective}, we introduce \iiql: an imitation learning algorithm that can learn intent-driven expert behavior from heterogeneous demonstrations. 

\subsection{\iiql: Intent-Driven Imitation Learner}
A na\"ive approach to solve Eq.~\ref{eq. main objective} is to first collect data of expert demonstrations $(s,a)$ and intent $(x)$ and then apply techniques for conventional imitation learning.
However, as human cognitive states cannot be readily measured using sensors, collecting intent data is non-trivial and resource-intensive in practice~\cite{bethel2007survey, heard2018survey, neubauer2020multimodal}.
Moreover, as detailed in Sec.~\ref{sec. oiql}, even in the case when intent data is available a na\"ive application of conventional imitation learning is inadvisable.
Hence, we derive an iterative factored approach to solve Eq.~\ref{eq. main objective} using the following two occupancy measures:
\begin{align}
    \rho_{\mathcal{N}}(s, a, x) &{\doteq} (1-\gamma) \Sigma_{t=0}^\infty \gamma^t p(s^t{=}s, a^t{=}a, x^t{=}x|\mathcal{N}, \mathcal{M})  \\
    \rho_{\mathcal{N}}(s, a, x^-) &{\doteq} (1-\gamma) \Sigma_{t=0}^\infty \gamma^t p(s^t{=}s, a^t{=}a, x^{t-1}{=}x^-|\mathcal{N}, \mathcal{M})  
\end{align}
We have the following proposition relating the three occupancy measures, which is proved in the Appendix.

\begin{proposition}
\label{thm. occupancy measure}
    Consider two arbitrary AMM models: $\mathcal{N}$ and $\mathcal{N}'$. Then,
    $\rho_{\mathcal{N}}(s, a, x) = \rho_{\mathcal{N}'}(s, a, x)$ and $\rho_{\mathcal{N}}(s, x, x^-) = \rho_{\mathcal{N}'}(s, x, x^-)$ if and only if $\rho_{\mathcal{N}}(s, a, x, x^-)=\rho_{\mathcal{N}'}(s, a, x, x^-)$.
\end{proposition}
Proposition \ref{thm. occupancy measure} implies that if the measures $\rho(sax)$ and $\rho(sxx^-)$ induced by the learner match with those induced by the expert, then $\mathcal{N} = \mathcal{N}_E$.
In other words, instead of directly optimizing $L_{saxx^-}$, one can alternatively solve the following factored sub-problems:
\begin{align}
    \min_{\mathcal{N}} L_{sax}(\mathcal{N}) &\doteq \min_{\mathcal{N}} D_f \left(\rho_{\mathcal{N}}(s, a, x) \middle|\middle| \rho_{\mathcal{N}_E} (s, a, x)\right)\label{eq. m-step pi objective}\\
    \min_{\mathcal{N}} L_{sxx^-}(\mathcal{N}) &\doteq \min_{\mathcal{N}} D_f \left(\rho_{\mathcal{N}}(s, x, x^-) \middle|\middle| \rho_{\mathcal{N}_E} (s, x, x^-)\right).\label{eq. m-step zeta objective}
\end{align}
Algorithm~\ref{alg. learner} provides an \textit{iterative} approach to solving these factored sub-problems via online imitation learning.
In line 2, \iiql initializes the estimate of expert's model as $\mathcal{N}_E \approx \mathcal{N}_{\theta, \phi} = (\pi_\theta$, $\zeta_\phi)$, where $\theta, \phi$ represent parameters of a known function (e.g. neural network).
We denote the iteration number as $n$ and the $n$-th estimate as $\mathcal{N}^n = (\pi^n, \zeta^n)$.
As the demonstrations generally do not include data of expert intent, line 4, performs intent inference using the current estimate of the expert model.
In lines 5-6, the algorithm augments its training data with on-policy experience.
The expert demonstrations and on-policy experience are then used to update the estimate of the expert model $(\pi, \zeta)$ (lines 7-8).
Due to its iterative nature, \iiql can also be viewed as an expectation-maximization (EM) algorithm in which the E-step corresponds to intent inference and M-step corresponds to model learning.
We detail \iiql's subroutines next.

\begin{algorithm}[t]
    \caption{\iiql: Intent-Driven Imitation Learner}
    \label{alg. learner}
    \begin{algorithmic}[1]
        \STATE \textbf{Input}: Data $D = \{\tau_{m}\}_{m=1}^d$ and $\{x_{m}^{0:h}\}_{m=1}^l$
        \STATE \textbf{Initialize}: $(\theta, \phi)$ where $\mathcal{N}_{\theta, \phi} = (\pi_\theta, \zeta_\phi)$
        \REPEAT
            \STATE E-step: Infer expert intent $\{x_m^{0:h}\}_{m>l}^d$ using $(\pi_\theta^n, \zeta_\phi^n)$ 
            \STATE Define $\tilde{D} \doteq D \cup \{\hat{x}_m^{0:h}\}_{m=1}^d$
            \STATE Generate trajectories $D_g = \{(s, x, a)^{0:h}\}$ with $(\pi_\theta^n, \zeta_\phi^n)$
            \STATE \pistep: Compute $\pi_\theta^{n+1}$ via Eq.~\ref{eq. m-step pi objective} using $\tilde{D}, D_g$ 
            \STATE \zetastep: Compute $\zeta_\phi^{n+1}$ via Eq.~\ref{eq. m-step zeta objective} using $\tilde{D}, D_g$
        \UNTIL{Convergence}
    \end{algorithmic}
\end{algorithm}

\subsubsection{E-step: Intent Inference}
Following algorithms for AMM learning~\cite{unhelkar2019learning, orlov2022factorial,seo2022semi}, we take a Bayesian approach to inferring values of expert intent.
Mathematically, to infer intent data $x^{0:h}$ for each demonstration $\tau$, \iiql computes the following maximum a posteriori (MAP) estimate for the posterior distribution:%
\begin{align}
    \hat{x}^{0:h} = \arg\max_{x^{0:h}}p(x^{0:h}|\tau, \mathcal{N}^n) \label{eq. e-step}
\end{align}%
using a variant of Viterbi algorithm introduced in \cite{viterbi1967error,unhelkar2019learning, jing2021adversarial}, which has a time complexity of $O(h|X|^2)$.
Observe that, using these estimates, one can augment the expert demonstrations $\tilde{D} \doteq D \cup \{\hat{x}_m^{0:h}\}_{m=1}^d$ to arrive at $(s,a,\hat{x})$-trajectories.
Further, these trajectories can then be used to compute the empirical estimates of expert's occupancy measures -- $\rho_E^n(s, a, x, x^-)$, $\rho_E^n(s, a, x)$, and $\rho_E^n(s, a, x^-)$ -- needed for the subsequent steps of the algorithm.

\subsubsection{\pistep: Learning Agent's Policy $\pi$}
\label{sec. pistep}
Assuming $\rho_E^n$ and $\zeta^{n}$ as fixed, in line 7, \iiql solves the sub-problem described in Eq.~\ref{eq. m-step pi objective} to update the agent's estimate of the expert policy $\pi_\theta$.
To derive the solution, we begin with an intent-augmented state representation, $u \doteq (s, x) \in S \times X$.
Using this representation, we define an intent-informed MDP $\tilde{\mathcal{M}} \doteq ((S \times X), A, \tilde{T}, \tilde{\mu}_0, \gamma)$, where  $\tilde{T}(u'|u, a) = T(s'|s, a) \zeta^n(x'|x, s)$ and $\tilde{\mu}_0(u)=\zeta(x|s, \#)\mu_0(s)$.
Observe that, for this intent-informed MDP, Eq.~\ref{eq. m-step pi objective} reduces to the conventional imitation learning problem of learning $\pi(a|u)$ by matching the occupancy measure $\rho(u,a) \equiv \rho(s,x,a)$.
Given this reduction, a variety of existing imitation learning algorithms can be used to solve Eq.~\ref{eq. m-step pi objective}.
Our implementation of \iiql uses a recent algorithm \iql to solve Eq.~\ref{eq. m-step pi objective} and update $\pi_\theta$.
This choice is motivated by \iql's state-of-the-art performance and stable training; unlike many related techniques, \iql and consequently \iiql do not rely on (potentially unstable) adversarial training.

\subsubsection{\zetastep: Learning Agent's Intent Dynamics $\zeta$}
Assuming $\rho_E^n$ and $\pi^{n+1}$ as fixed, in line 8, \iiql updates $\zeta_\phi$ by solving Eq.~\ref{eq. m-step zeta objective}.
To derive the solution, we utilize an alternate intent-augmented state representation $v\doteq (s, x^-) \in S\times X^+$.
Using this representation and modeling the set of intents $X$ as agent's macro-actions, we define the intent-transition MDP $\bar{\mathcal{M}} = ((S\times X^+), X, \bar{T}, \bar{\mu}_0, \gamma)$, where the transition model of $v$ is defined as $\bar{T}(v'|v, x) = \sum_a \pi^{n+1}(a|s, x) T(s'|s, a)$ and $\bar{\mu}_0(v) = \mu_0(s)$ represents the initial distribution of $v$.
Observe that $\zeta(x|v)$ corresponds to the policy of the intent-transition MDP $\bar{\mathcal{M}}$, which describes the agent's approach to selecting its intent $x$ given the task state and prior intent $v = (s,x^-)$.
Hence, for the intent-transition MDP, Eq.~\ref{eq. m-step zeta objective} too reduces to the conventional imitation learning problem of learning $\zeta(x|v)$ by matching the occupancy measure $\rho(v,x) \equiv \rho(s,x,x^-)$.
Consequently, similar to \pistep, a variety of existing imitation learning algorithms can be used to solve Eq.~\ref{eq. m-step zeta objective}.
For reasons stated in Sec.~\ref{sec. pistep}, our implementation of \iiql uses \iql to also solve Eq.~\ref{eq. m-step zeta objective} and update agent's estimate of the expert's intent dynamics $\zeta_\phi$.

\subsubsection{Implementation Considerations}
\label{sec. implementation considerations}
As suggested by our analysis presented next, we opt for small learning rates while updating the model parameters.
This leads to small incremental updates to the parameters $\theta$ and $\phi$ in the \pistep and \zetastep.
Further, to enhance learning performance, we conduct multiple iterations of \pistep and \zetastep (lines 7-8) in every loop of Algorithm~\ref{alg. learner} after the subsequent E-step (line 4) in our implementation. For detailed insights and other considerations, please refer to the Appendix.

\subsection{Sufficient Conditions for Convergence}
\label{sec. sufficient conditions}
Proposition \ref{thm. occupancy measure} establishes that $\mathcal{N} = \mathcal{N}_E$ when both $L_{sax} = 0$ and $L_{sxx^-}=0$.
However, due to the iterative nature of Algorithm \ref{alg. learner}, this alone does not ensure the convergence to the expert model.
Minimizing $L_{sax}$ in \pistep might inadvertently increase $L_{sxx^-}$, and the opposite could occur in \zetastep.
To address this, we derive the conditions under which  Algorithm~\ref{alg. learner} monotonically reduces the primary loss function (Eq.~\ref{eq. main objective}) and, thus, asymptotically converges to a sound estimate of the expert's behavioral model.

Consider the following approximations of the expert occupancy measure ${\rho}_E(s, a, x, x^-)$, which are computed using the $n$-th estimate of the expert model and factored occupancy measures:
\begin{align}
   \check{\rho}_E^n(\cdot) = \check{\rho}_E^n(s, a, x, x^-) &\doteq \sum_{s^-, a^-}\pi^n(\cdot) \zeta^n(\cdot) T(\cdot) \rho_E^n(s^-, a^-, x^-) \label{eq:check-rho} \\
   \grave{\rho}_{E}^n(\cdot) = \grave{\rho}_{E}^n(s, a, x, x^-) &\doteq \pi^n(a|s,x) \rho_E^n(s, x, x^-)
   \label{eq:grave-rho}
\end{align}%
For these approximations, we have the following two lemmas.\footnote{Please refer to the Appendix for detailed proofs.}
\begin{lemma}
    \label{thm. pi objective}
    Define $|\Delta(\theta,\theta_n)| \doteq \epsilon$.
    If $\epsilon$ is sufficiently small, then
    \begin{equation*}
        D_f\left(\rho_{\pi_\theta}(s, a, x)\middle|\middle|\rho_E^n(s, a, x)\right) = D_f\left(\rho_{\pi_\theta, \zeta^n}(s, a, x, x^-)\middle|\middle|\check{\rho}_E^n(\cdot)\right)
    \end{equation*}
\end{lemma}
\begin{lemma}
\label{thm. zeta objective}
    Define $|\Delta(\theta_{n+1},\theta_n)| \doteq \epsilon$.
    If $\epsilon$ is sufficiently small, then
    \begin{equation*}
        D_f\left(\rho_{\zeta_{\phi}}(s, x, x^-)\middle|\middle|\rho_E^n(s, x, x^-)\right)= D_f\left(\rho_{\pi^{n+1}, \zeta_{\phi}}(s, a, x, x^-)\middle|\middle|\grave{\rho}_{E}^n(\cdot)\right)
    \end{equation*}
\end{lemma}

Lemma~\ref{thm. pi objective} implies that, if the  update to policy parameters in \pistep is sufficiently small,\footnote{In practical implementations, sufficiently small updates to the policy parameters can be realized through appropriate choice of the learning rate hyperparameter.} then the value of sub-objective $L_{sax}$ approximately equals the overall objective $L_{saxx^-}$.
Here, the approximation arises from the use of $n$-th estimates of expert model to approximate the expert occupancy measures $\rho_E(s,a,x,x^-) \approx \check{\rho}_E^n(\cdot)$.
Under similar conditions and the approximation $\rho_E(s,a,x,x^-) \approx \grave{\rho}_E^n(\cdot)$, Lemma~\ref{thm. zeta objective} implies that the value of sub-objective $L_{sxx^-}$ equals the overall objective $L_{saxx^-}$. 
Thus, under the given conditions, a decrease in $L_{sax}$ and $L_{sxx^-}$ will also decrease $L_{saxx^-}$.
The proof of these lemmas rely on first-order approximations with respect to the policy parameters and convexity of the function used to compute the $f$-divergence.

Given these lemmas, we can now proceed to stating the sufficient conditions for convergence of \iiql.
Consider a third approximation of the expert occupancy measure ${\rho}_E(s, a, x, x^-)$ computed using the Viterbi intent estimates derived in the E-step as follows:
\begin{align}
   {\rho}_E^n(\cdot) = \rho_E^n(s,a,x,x^-) \doteq \rho_E(s, a)p(x, x^-|s, a, \mathcal{N}^n)
\end{align}%
Further, denote the empirical estimate of the original objective of Eq.~\ref{eq. main objective} at the $n$-th iteration as 
\begin{equation}
    L_{saxx^-} \approx L_{saxx^-}^n {\doteq}  D_f \left(\rho_{\mathcal{N}}(s, a, x, x^-) \middle|\middle| \rho_{E}^n (s, a, x, x^-)\right)
\end{equation}
Then, we have the following theorem for the convergence of \iiql.
\begin{theorem}\label{thm. convergence}
    If (1) $\rho_E^n \approx \check{\rho}_E^n \approx \grave{\rho}_E^n$; and (2) $|\Delta(\theta_{n+1},\theta_n)|$ is sufficiently small, then $L_{saxx^-}^{n+1}\leq L_{saxx^-}^n$.
\end{theorem}
Note that, while important, Theorem~\ref{thm. convergence} only guarantees convergence to a local optima computed with respect to the approximate objective $\lim_{n\to\infty} L_{saxx^-}^n$.
A stronger result can be achieved for the supervised case of the intent-driven imitation learning, in which labels of expert intent available for all expert demonstrations.\footnote{For the supervised case, the E-step of Algorithm~\ref{alg. learner} is not required, and offline pre-training can greatly help.
}
As evidenced in our experiments, however, the empirical results are more encouraging.
\iiql, even with no or small supervision, converges to the expert behavior in a variety of problems.

\subsection{\oiql: A joint variant of \iiql}
\label{sec. oiql}
Given the inferred values of expert intent, \iiql factors the original problem (Eq.~\ref{eq. main objective}) into two sub-problems (Eq.~\ref{eq. m-step pi objective} and Eq.~\ref{eq. m-step zeta objective}) which are each solved separately.
As the analysis of its convergence relies on showing that optimizing the two factored objectives is equivalent to optimizing the original joint objective $L_{saxx^-}$, a natural question arises: \textit{why not directly optimize the joint objective?}
To answer this question, we provide an alternate approach that indeed uses the inferred intent values to directly optimize the original joint objective $L_{saxx^-}$.
To derive this approach, consider the joint MDP model, with state defined as $w \doteq (s, x^-) \in S\times X^+$ and actions as $a_w \doteq (a, x) \in A\times X$.
The policy for this joint MDP corresponds to a joint  estimate of the expert policy and intent dynamics
\begin{equation}
 \tilde{\pi}(a_w|w) \doteq \tilde{\pi}(a,x|s,x^-) = \pi(a|s, x) \zeta(x|s, x^-)   
\end{equation}
Further, given the inferred values of $(x,x^-)$ computed via the E-step and the aforementioned MDP, Eq.~\ref{eq. main objective} reduces to the conventional imitation learning problem of learning $\tilde{\pi}(a_w|w)$ by matching the occupancy measure $\rho(w,a_w) \equiv \rho(s,a,x,x^-)$.
Given this reduction, similar to the factored \pistep and \zetastep, we can now apply existing techniques to solve the original problem.
For instance, we can apply \iql to first learn $Q(w, a_w) = Q(s, x^-, a, x)$ from the expert demonstrations and online experience.
Next, using the learnt $Q$-values we can then recover the expert model ($\pi$ and $\zeta$ ) by optimizing the following soft actor-critic (SAC) objective~\cite{haarnoja2018soft}:
\begin{align}
    &J(\theta, \phi) = E_{(s, x^-) \sim \tilde{D}}\left[D_{KL}\left( \pi_\theta \cdot \zeta_\phi  \middle|  \middle|\frac{\exp(Q(s, x^-, a, x))}{Z}\right) \right] \label{eq. joint sac}
\end{align}
We refer to this unfactored approach, which replaces the two  \pistep and \zetastep with a joint M-step, as \iiql-Joint or \oiql in short.
This approach is technically sound.
However, as learning the joint function does not leverage the inherent factored structure of the expert behavior, we anticipate its learning performance and interpretability to be subpar than \iiql.
In Sec.~\ref{sec. experiments}, we evaluate this hypothesis by empirically comparing \iiql and \oiql.

\begin{figure*}[t]
  \centering
  \newcommand\gap{0.18}
  \begin{subfigure}[t]{\gap\linewidth}
      \centering
      \includegraphics[width=\textwidth, frame]{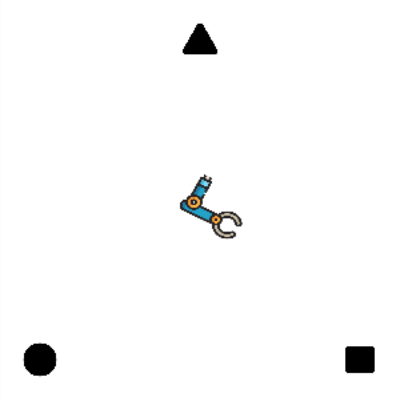}
      \caption{\twodim}
      \label{fig. multigoal}
  \end{subfigure}
  \hfill
    \begin{subfigure}[t]{\gap\linewidth}
      \centering
      \includegraphics[width=\textwidth,height=3.4cm]{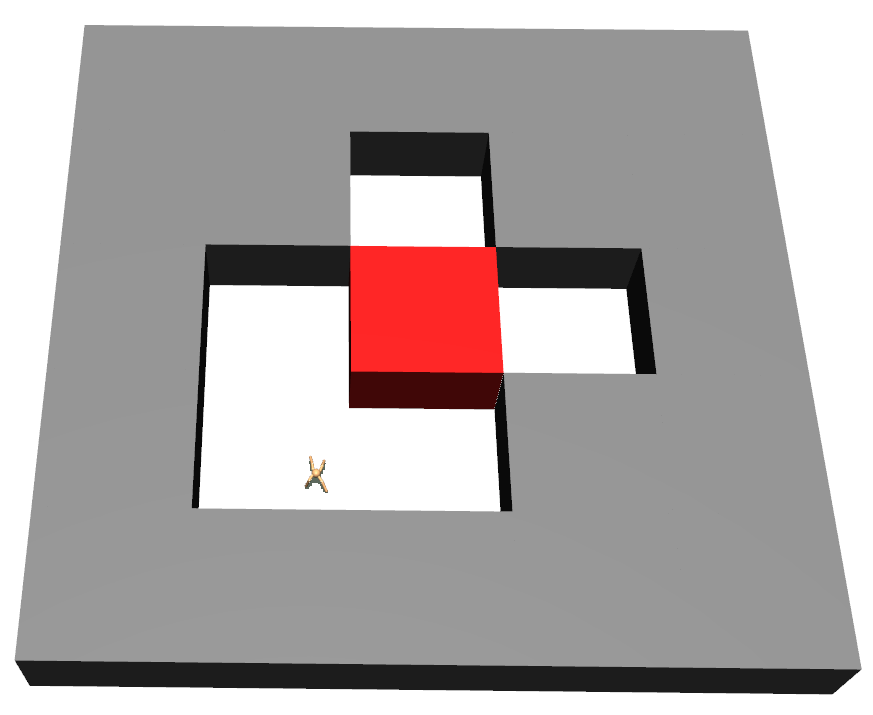}
      \caption{\texttt{AntPush}}
      \label{fig. antpush}
  \end{subfigure}
  \hfill
  \begin{subfigure}[t]{\gap\linewidth}
      \centering
      \includegraphics[width=\textwidth]{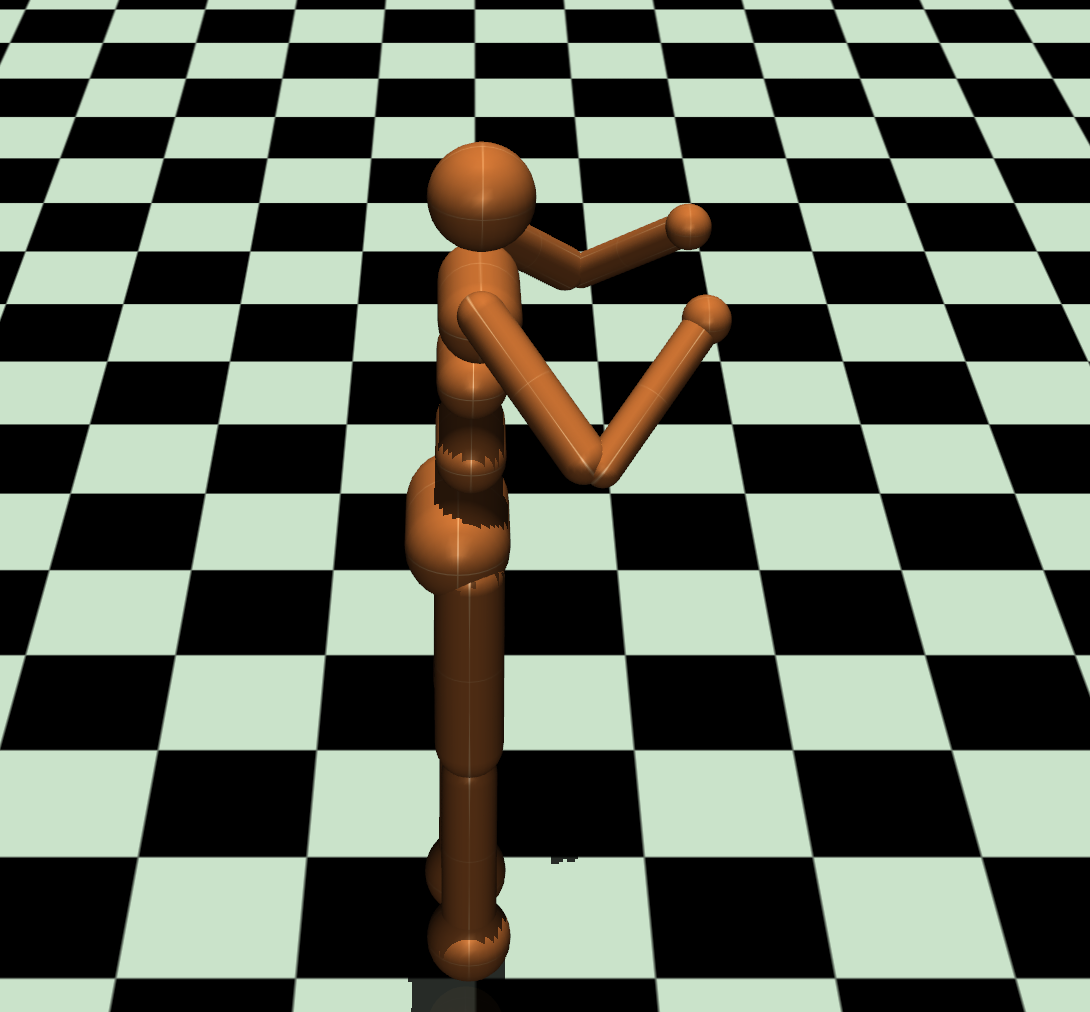}
      \caption{\texttt{Humanoid}}
      \label{fig: humanoid}
  \end{subfigure}
  \hfill
  \begin{subfigure}[t]{\gap\linewidth}
      \centering
      \includegraphics[width=\textwidth, frame]{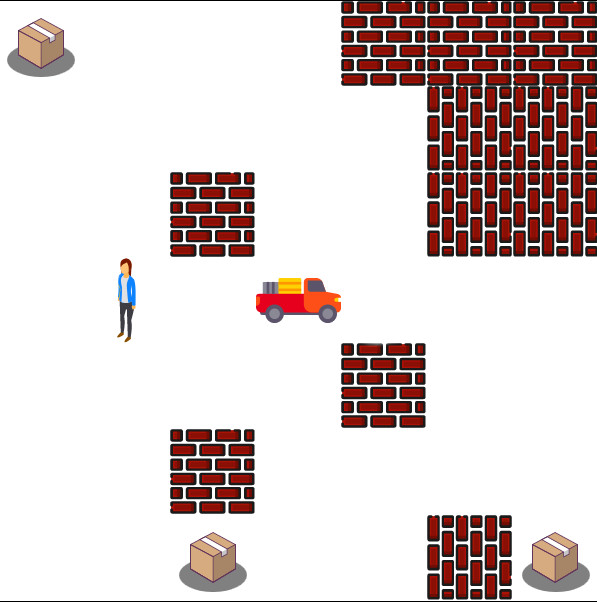}
      \caption{\single}
      \label{fig: one mover}
  \end{subfigure}
    \hfill
  \begin{subfigure}[t]{\gap\linewidth}
      \centering
      \includegraphics[width=\textwidth, frame]{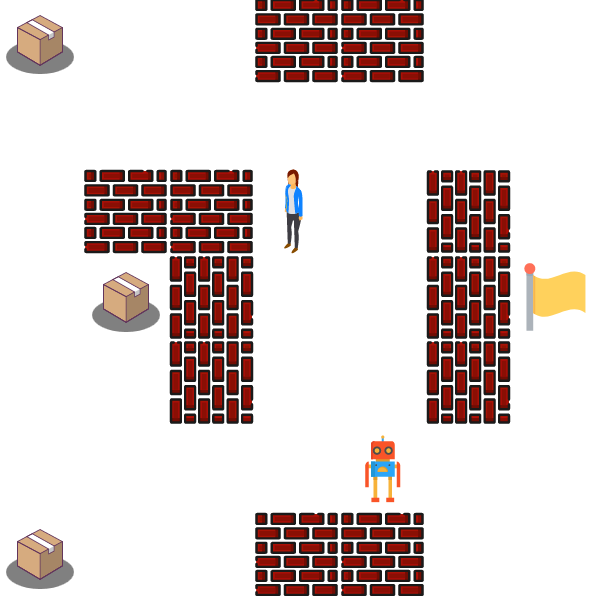}
      \caption{\movers}
      \label{fig: two mover}
  \end{subfigure}
  \captionsetup{subrefformat=parens}
  \caption{Visual illustrations of Experimental Domains.}
  \label{fig: mujoco}
\end{figure*}

\subsection{Relation to Existing Works}
\label{sec. related works}
We conclude this section by discussing the relationship of \iiql to relevant human modeling and imitation learning techniques that also consider latent decision-making factors~\cite{hiatt2017human, albrecht2018autonomous,arora2021survey,osa2018algorithmic}.

\subsubsection{Human Behavior Modeling}
A prevailing motif in human modeling research is the significance of mental (or latent) states, such as intent, expertise, and trust in the robot~\cite{thomaz2016computational,orlov2024rw4t,qian2024measuring,yang2017evaluating,qian2022evaluating}.
Techniques that incorporate these latent states generally outperform those that exclude them~\cite{choi2012nonparametric, schmerling2018multimodal,ivanovic2018generative, li2017infogail,hausman2017multi,wang2022co,xie2021learning,beliaev2022imitation}.
For instance, explicit modeling of human intent has been used to learn more effective policies for human-robot collaboration~\cite{xie2021learning}.
Similarly, extensions of Generative Adversarial Imitation Learning (GAIL) that incorporate demonstrator's strategy code as an latent feature tend to outperform GAIL~\cite{li2017infogail,hausman2017multi,wang2022co, ho2016generative}.
While these methods learn human behavior with diverse intents, unlike \iiql, they assume that the intent remains static during an episode.

In contrast, Agent Markov Model (AMM) and its variants consider both the expert's latent states and its temporal dynamics~\cite{unhelkar2019learning,orlov2022factorial,seo2022semi}.
Thus, the model can capture diverse and dynamically-changing intents representative of human behavior.
However, existing learning methods for AMM utilize variational inference and tabular representations, which imposes strong assumptions on the target distributions and cannot scale up to complex tasks.
\iiql does not make such assumptions and thus is capable of learning expert behavior in tasks with large or continuous state spaces.

\subsubsection{Hierarchical Imitation Learning}
As intents typically shape expert behavior at a macro level, the problem of learning intent-driven behavior  shares similarities with hierarchical imitation learning~\cite{le2018hierarchical,kipf2019compile,bogert2022hierarchical,zhang2021provable,jain2024godice}.\footnote{For instance, in our problem setting, intent can be viewed as a one-step option~\cite{sutton1999between}. However, in contrast to the options framework, intent is neither associated with an option termination condition nor assumes temporal abstraction.}
\citet{le2018hierarchical,kipf2019compile,zhang2021provable} propose hierarchical policy learning algorithms based on behavior cloning. 
These methods, however, do not utilize task dynamics and are prone to compounding errors~\cite{tu2022sample,ross2011reduction}. 
To alleviate the problem of compounding errors, several GAIL-based approaches have been proposed to learn hierarchical policies~\cite{sharma2018directed,lee2020learning,jing2021adversarial,chen2023option}. 
Among these, \ogail that builds upon the options framework and utilizes an EM approach is most relevant to our work~\cite{jing2021adversarial,zhang2019dac,sutton1999between}.
However, unlike our approach, \ogail relies on adversarial training and thus is prone to unstable gradients~\cite{arenz2020non,jiang2018computation}.
We posit, through the use of a non-adversarial approach, \iiql will result in stable training and superior performance~\cite{garg2021iq}.
We evaluate this hypothesis in the following section.

\begin{table}[t]
\caption{Key Characteristics of Experimental Domains}
\label{table. domains}
\begin{center}
\begin{tabular}{@{}lcc@{\hskip 12pt}cc@{}} \toprule
                 & \multicolumn{2}{c}{\bf State Space}                  
                 & \multicolumn{2}{c}{\bf Expert} \\ \cmidrule(lr){2-3} \cmidrule(lr){4-5}
\textbf{Domain}           & Type           & $|S|$ or $n_s$ & Intent       & Human       \\ \midrule
\twodimshort-n suite   & Continuous  &    2       & Yes          &             \\
\texttt{AntPush}          & Continuous     &     107       & Maybe        &             \\
\texttt{Mujoco} suite      & Continuous     &    376     &              &             \\
\single          & Discrete       &      $\approx1k$      & Yes          &             \\
\movers          & Discrete       &    $\approx40k$     & Yes          & Yes         \\
\bottomrule      
\end{tabular}
\end{center}
\end{table}

\section{Experiments}
\label{sec. experiments}
We evaluate \iiql on multiple domains and benchmark its ability to successfully complete tasks, infer expert intent, and generate diverse interpretable behaviors.
Through this empirical investigation, we evaluate, in terms of performance,
\begin{enumerate}
    \item[(Q1)] How does \iiql compare to conventional IL techniques when experts exhibit intent-driven behavior?
    \item[(Q2)] How does \iiql compare to conventional IL techniques when experts \textit{do not} exhibit intent-driven behavior?
    \item[(Q3)] How does \iiql compare to hierarchical IL techniques?
    \item[(Q4)] How does \iiql compare to its variant \oiql?
\end{enumerate}

\subsection{Domains and Expert Demonstrations}
\begin{table*}[ht]
\caption{Cumulative task reward (averaged over three  experiment trials) for each method.}
\label{table. task reward results}
\begin{center}
\setlength\tabcolsep{3pt} 
\begin{tabular}{lllllll:lll} \toprule
\multicolumn{1}{c}{\bf Domain} & \multicolumn{1}{c}{\bf Expert} & \multicolumn{1}{c}{\bf BC} & \multicolumn{1}{c}{\bf \iqlshort} & \multicolumn{1}{c}{\bf \ogailshort} & \multicolumn{1}{c}{\bf \oiqlshort} & \multicolumn{1}{c:}{\bf \iiql} & \multicolumn{1}{c}{\bf \ogailsemi} & \multicolumn{1}{c}{\bf \oiqlsemi} & \multicolumn{1}{c}{\bf \iiqlsemi} \\
\midrule
\twodimshort-$2$ & \mustd{15.19}{1.0} & \mustd{-8.69}{2.3} & \mustd{-2.61}{8.4} & \mustd{-10.51}{2.4} & \mustd{1.08}{6.7} & \textbf{\mustd{16.16}{0.3}} & \mustd{-9.10}{5.2} & \mustd{1.55}{1.2} & \textbf{\mustd{16.20}{0.2}} \\
\twodimshort-$3$ & \mustd{22.32}{0.9} & \mustd{-7.08}{1.5} & \mustd{6.06}{14.8} & \mustd{-11.25}{2.2} & \mustd{-4.58}{1.4} & \textbf{\mustd{20.64}{2.2}} & \mustd{-4.89}{5.9} & \mustd{-3.33}{0.7} & \textbf{\mustd{22.25}{0.8}} \\
\twodimshort-$4$ & \mustd{29.14}{1.4} & \mustd{-10.4}{0.7} & \mustd{-8.67}{2.3} & \mustd{-13.33}{3.6} & \mustd{-4.17}{1.4} & \textbf{\mustd{13.63}{2.1}} & \mustd{-13.75}{3.3} & \mustd{-4.17}{2.6} & \textbf{\mustd{24.61}{1.7}} \\
\twodimshort-$5$ & \mustd{40.09}{1.4} & \mustd{-6.25}{1.3} & \mustd{-10.92}{1.8} & \mustd{-12.92}{2.6} & \mustd{4.17}{4.4} & \textbf{\mustd{27.88}{1.8}} & \mustd{-12.08}{5.1} & \mustd{6.27}{2.5} & \textbf{\mustd{32.84}{0.7}} \\ 
\midrule
\single          & \mustd{-46.8}{4} & \mustd{-200.0}{0} & \mustd{-45.7}{3} & \mustd{-200.0}{0} & \mustd{-91.3}{17} &\textbf{ \mustd{-45.5}{1}} & \mustd{-200.0}{0} & \mustd{-70.6}{32} & \textbf{\mustd{-46.0}{2}} \\
\movers          & \mustd{-67.6}{6} & \mustd{-200.0}{0} &\textbf{ \mustd{-65.6}{0}} & \mustd{-200.0}{0} & \mustd{-152.3}{16} & \mustd{-68.3}{3} & \mustd{-200.0}{0} & \mustd{-155.2}{28} & \textbf{\mustd{-68.2}{1}} \\
\midrule
\texttt{Hopper}           & \mustd{3533}{39} & \mustd{861}{558} & \mustd{3523}{22} & \mustd{3279}{426} & \textbf{\mustd{3558}{40}} & \mustd{3523}{21} & \multicolumn{3}{c}{\multirow{6}{*}{N/A}} \\
\textonehalf \texttt{Cheetah} & \mustd{5098}{62} & \mustd{4320}{314} & \mustd{5092}{31} & \mustd{2801}{923} & \textbf{\mustd{5136}{37}} & \mustd{5098}{66} & \\
\texttt{Walker2d}      & \mustd{5274}{53} & \mustd{3065}{800} & \mustd{5185}{43} & \mustd{4178}{1610} & \mustd{5179}{31} & \textbf{\mustd{5249}{19}}  &   \\
\texttt{Ant}           & \mustd{4700}{80} & \mustd{4179}{175} & \mustd{4606}{38} & \mustd{4529}{59} & \textbf{\mustd{4636}{27}} & \mustd{4580}{30}  &    \\
\texttt{Humanoid}      & \mustd{5313}{210} & \mustd{496}{47} & \mustd{5350}{58} & \mustd{458}{43} & \mustd{5298}{22} & \textbf{\mustd{5370}{133}}  &    \\
\texttt{AntPush}       & \mustd{116.6}{14} & \mustd{90.2}{13} & \mustd{116.6}{1} & \mustd{98.9}{16} & \mustd{116.6}{2} & \textbf{\mustd{117.1}{0} }  &   \\
\bottomrule
\end{tabular}
\end{center}
\end{table*}

As summarized in Table~\ref{table. domains}, we consider both discrete and continuous MDPs, low- and high-dimensional state spaces, intent-driven and intent-agnostic behaviors, and human and artificial experts.
In Table~\ref{table. domains}, we list the size of the state space $|S|$ for discrete domains and number of state features $n_s$ for continuous domains.
For the \twodim-n and \mujoco suites, we report these values for the most complex task of the suite.

\subsubsection{\twodim-$n$ (\twodimshort-$n$).} As introduced in Sec.~\ref{sec. motivating scenario}, this domain models an agent tasked with visiting $n$ landmarks (goals) in an empty environment of size $5\times5$.
Figure~\ref{fig. multigoal} depicts this domain with $n=3$.
No constraint is placed on the visitation order of landmarks, which is decided based on expert's intent.
Both state and action are 2-dimensional vectors, denoting the position and velocity of the agent.
The agent receives a reward of $10$ upon reaching a previously unvisited landmark and a penalty of $0.1$ every time step.
We synthetically generate 100 demonstrations (50/50 train/test split) for this task by hand-crafting an expert model.
The hand-crafted expert first selects an arbitrary unvisited landmark and then performs goal-oriented motion to visit it.
Observe that the state alone is insufficient to predict expert behavior, as it depends on both its location and intent (i.e., which landmark the expert plans to visit next).
We consider four versions of this task, by varying $n$ from two to five.

\subsubsection{\single.}
This domain, depicted in Figure~\ref{fig: one mover},  models a single-agent version of the benchmark domain \texttt{Movers} introduced by \citet{seo2023automated}.
It simulates a $7\times 7$ discrete grid world, in which the agent is task with moving boxes to the truck.
The state denotes task information (agent and box locations) but not the agent intent (i.e., which box it plans to pick or drop next). 
We assume an expert may have one of four different intents: going to one of the three boxes or a truck).
Similar to \twodim-$n$, we synthetically generate 100 demonstrations (50/50 train/test split) for this task by hand-crafting an expert model.

\subsubsection{\mujoco suite.} 
To evaluate \iiql in domains that may not involve diverse or time-varying intents, we consider the following benchmark \mujoco environments: \texttt{Hopper-v2}, \texttt{HalfCheetah-v2}, \texttt{Walker2d-v2}, \texttt{Ant-v2}, \texttt{Humanoid-v2}, and \texttt{AntPush-v0} \cite{todorov2012mujoco}.
We use the version of \texttt{AntPush-v0} introduced by \citet{jing2021adversarial} and their expert dataset to facilitate a direct comparison between the performance of \iiql and \ogail.
For the remaining tasks, we employ the dataset made available by \citet{garg2021iq}.
We use demonstrations of 1k, 5k, and 50k time steps for  \texttt{Hopper-v2}, \texttt{Walker2d-v2}, and \texttt{AntPush-v0}, respectively, the same as \citet{jing2021adversarial}.
For other tasks, we use five trajectories (5k time steps).
For all tasks, we set the number of intents to four following \citet{jing2021adversarial}.

\subsubsection{\movers}
Modeling human behavior and recognizing their intent are particularly important for human-AI collaboration.
Hence, we also utilize \movers: a human-AI collaboration setting introduced by \citet{seo2022semi}.
This domain models a human-agent team that needs to work together to move boxes to a truck.
Originally a multi-agent problem, we adapt it to a single-agent context by treating the robot teammate as an environmental component. 
We utilize 66 human demonstrations, 44 for training and 22 for testing.

\subsection{Baselines and Metrics}
\subsubsection{Baselines} 
We benchmark \iiql and its variant \oiql against Behavioral Cloning (BC), \iql (\iqlshort), and \ogail (\ogailshort) \cite{pomerleau1988alvinn, garg2021iq,jing2021adversarial}.
BC is a simple yet effective IL technique \cite{li2022rethinking}.
\iql and \ogail  represent the state of the art in conventional and hierarchical IL, respectively.
For the intent-aware techniques, we also benchmark the performance of their semi-supervised variants (denoted as \iiqlsemi, \oiqlsemi, \ogailsemi), which additionally receive ground truth labels of expert intent for 20\% of the trajectories.

\subsubsection{Implementation Details}
We utilize a multi-layer perceptron (MLP) with two hidden layers, each containing $128$ nodes, to implement all neural networks of \iiql; this includes Q-networks for $\zeta$ and $\pi$, and the actor network for $\pi$ in the case of continuous domains.
Other hyperparameters are selected via grid search.
The policy network for BC employs an MLP with two hidden layers of $256$ nodes each.
For implementing \iql and \ogail, we adopt the neural network structures and hyperparameters proposed in their original papers \cite{garg2021iq, jing2021adversarial}. 
\oiql shares the same architecture as \iql.
All online imitation learning methods were trained until convergence, capped at 6M exploration steps.
Behavior Cloning was trained for 10k updates, using a batch size of 256.
Please refer to the Appendix for additional implementation details.

\subsubsection{Metrics}
We consider three metrics: the cumulative task reward, accuracy of intent inference, and the diversity of behaviors generated by the learned model.
To characterize the learning curve, cumulative reward is computed every 20k exploration steps as an average over $8$ evaluation episodes.
Table~\ref{table. task reward results} reports the maximum of this learning curve.
We determine the accuracy of intent inference by computing the fraction of time steps wherein the algorithm accurately identifies the intent, relative to the total steps in the test set.
The range and interpretability of behaviors exhibited by the learned model is qualitatively assessed through a visual inspection of its generated trajectories.

\subsection{Results}

\subsubsection{Task Performance: Intent-Driven Experts} 
First, we consider the performance of \iiql in domains where experts exhibit diverse intent-driven behaviors: namely, \twodim, \single, and \movers.
As shown in Table \ref{table. task reward results}, \textit{\iiql either outperforms or performs comparably to the baselines} in these domains.
Further, in domains where the performance is comparable to baselines, both \iiql and the best performing baseline perform as well as the expert.

Towards Q1, we observe that \iql (a conventional IL technique) performs as well as the expert in some intent-driven tasks but not in others.
Its task performance is especially poor in \twodim domain, where expert behaviors are not only diverse but also contradictory.
For instance, based on their intent, different experts may select opposite actions at the same state.
As \iql does not model intent it is unable to resolve this ambiguity and thus leads to subpar performance.
In contrast, \iiql is still able to mimic expert behavior and achieves a high task reward, associating contradictory behaviors to different intents.
Moreover, even in domains where \iql exhibits expert-level performance, the learnt model cannot be used to infer the actor's intent or generate diverse behaviors.

Among intent-aware algorithms considered in Q3 and Q4, \iiql outperforms \ogail and \oiql.
\ogail particularly performs poorly in \single and \movers.
We expect that this occurs due to the large input dimension for its discriminator, caused by the conversion of a discrete state into a corresponding one-hot encoding. 
Such high-dimensional inputs hinder the performance of the discriminator, which is crucial for GAIL-like algorithms to work, by making it challenging for a neural network to differentiate between them.
\oiql also shows poor performance in discrete domains, presumably due to Gumbel-Softmax reparameterization \cite{jang2016categorical} of Categorical distribution required for backpropagation of gradients when optimizing the SAC objective.
As mentioned in Sec. \ref{sec. implementation considerations}, since \iiql learns a separate $Q(s,a,x)$-function, the policy for discrete action can be directly derived from the learnt $Q$, removing the additional optimization for the SAC objective.
We also observe that semi-supervision of intents leads to faster convergence and improves the performance of \iiql (cf. Table~\ref{table. task reward results}: \iiql and \iiqlsemi).
While \ogail and \oiql also benefit from this additional information, the improvement is the most noticeable for \iiql. 

\subsubsection{Task Performance: Intent-Agnostic Experts}
For the domains where demonstrations do not include diverse or intent-driven expert behaviors, \iql, \oiql, and \iiql all show comparable performance and achieve the near-expert task reward.
Towards Q2, this demonstrates that \iiql does not overfit to the intent-driven model and does not collapse when demonstrations are homogeneous.
Also, \iiql's performance in the \mujoco suite ensures that it can handle expert's demonstrations in complex and continuous domains, which is not possible with recent AMM-learning algorithms~\cite{unhelkar2019learning,seo2022semi,orlov2022factorial}.
In addition, compared to \ogail and \oiql, \iiql achieves a higher task reward and show lower variance. Taking into account that \oiql and \ogail utilize identical objectives Eq.~\ref{eq. main objective}, this results demonstrate the advantage of non-generative adversarial approach for achieving consistent training.

\subsubsection{Intent Inference}
\begin{table}[t]
\caption{Accuracy of Intent Inference}%
\label{table. inference}
\begin{center}
\setlength\tabcolsep{4.5pt} 
\begin{tabular}{lclll} \toprule
\multicolumn{1}{l}{\bf Domain} & \multicolumn{1}{c}{\bf Random} & \multicolumn{1}{c}{\bf \ogailsemi} & \multicolumn{1}{c}{\bf \oiqlsemi} & \multicolumn{1}{c}{\bf \iiqlsemi}\\ \midrule
\twodimshort-$3$ & $\approx0.33$ & \mustd{0.44}{0.1} & \mustd{0.64}{0.1}  & \textbf{\mustd{0.93}{0.0}}  \\
\twodimshort-$5$ & $\approx0.20$ & \mustd{0.20}{0.0} & \mustd{0.53}{0.2}  & \textbf{\mustd{0.83}{0.1}}  \\
\single & $\approx0.25$ & \mustd{0.38}{0.1} & \mustd{0.32}{0.2} &\textbf{\mustd{0.62}{0.0}} \\
\movers & $\approx0.20$ & \mustd{0.38}{0.3} & \mustd{0.48}{0.1} & \textbf{\mustd{0.60}{0.1}} \\
\bottomrule
\end{tabular}
\end{center}
\end{table}
A high reward demonstrates that the algorithm is capable of finding one (near-)optimal behavior. However, it does not imply that the algorithm is adept in capturing intents or diversity observed in expert behavior.
Hence, we benchmark the intent inference performance computed using the learnt models in Table \ref{table. inference}.
To associate the expert and learnt intent, we train models with 20\% supervision of intent.
Notice that, since \iql and Behavior Cloning learn a single behavior, models learnt using them cannot be used for intent inference.
We observe that \iiql outperforms \ogail, \oiql, and random guess in all domains.

\begin{figure}[t]
  \def\figwid{.27}
  \centering
  \begin{subfigure}[t]{\linewidth}
      \centering
      \includegraphics[width=\figwid\linewidth, frame]{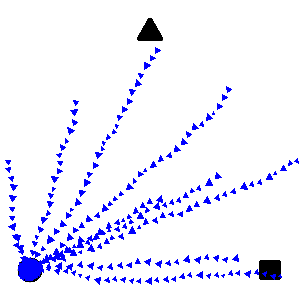}\hspace{1ex}
      \includegraphics[width=\figwid\linewidth, frame]{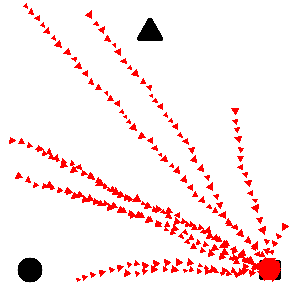}\hspace{1ex}
      \includegraphics[width=\figwid\linewidth, frame]{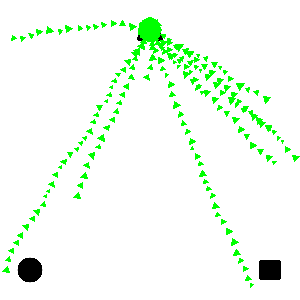}
      \caption{Expert}
  \end{subfigure}
  \begin{subfigure}[t]{\linewidth}
      \centering
      \includegraphics[width=\figwid\linewidth, frame]{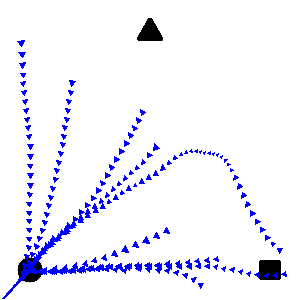}\hspace{1ex}
      \includegraphics[width=\figwid\linewidth, frame]{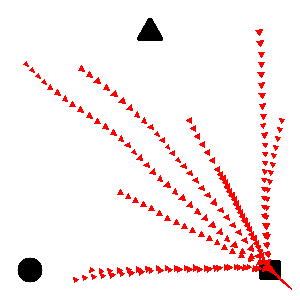}\hspace{1ex}
      \includegraphics[width=\figwid\linewidth, frame]{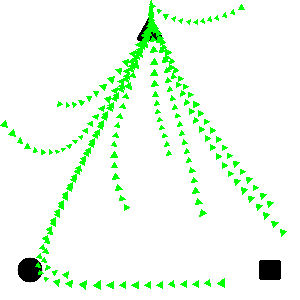}
      \caption{\iiql}
  \end{subfigure}
  \begin{subfigure}[t]{\linewidth}
      \centering
      \includegraphics[width=\figwid\linewidth, frame]{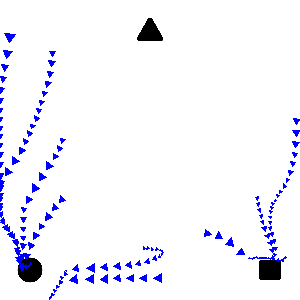}\hspace{1ex}
      \includegraphics[width=\figwid\linewidth, frame]{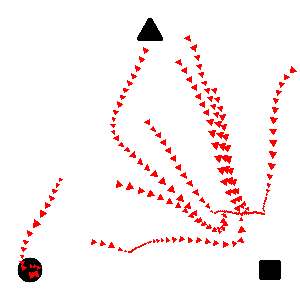}\hspace{1ex}
      \includegraphics[width=\figwid\linewidth, frame]{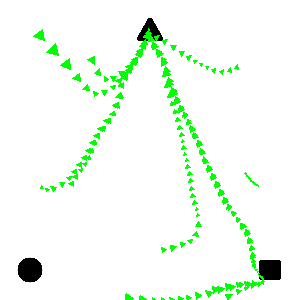}
      \caption{\oiql}
  \end{subfigure}
  \begin{subfigure}[t]{\linewidth}
      \centering
      \includegraphics[width=\figwid\linewidth, frame]{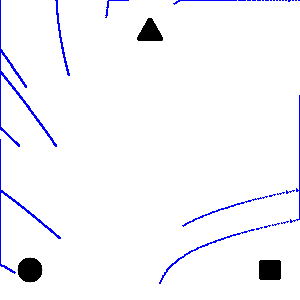}\hspace{1ex}
      \includegraphics[width=\figwid\linewidth, frame]{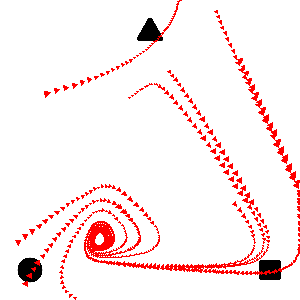}\hspace{1ex}
      \includegraphics[width=\figwid\linewidth, frame]{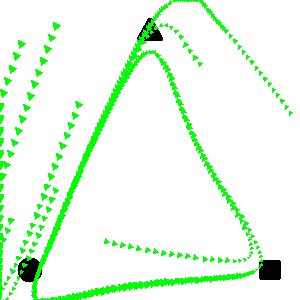}
      \caption{\ogail}
  \end{subfigure}
  \captionsetup{subrefformat=parens}
  \caption{\twodim-$3$ trajectories generated by the expert and learnt models according to different intents.}
  \label{fig: paths}
\end{figure}
\subsubsection{Diversity and Interpretability of Generated Behaviors}
Finally, in Figure~\ref{fig: paths}, we visualize and qualitatively inspect the diverse intent-driven behaviors learnt by various algorithms for the \twodim-3 domain.
Recall that in this domain each intent-driven behavior corresponds to reaching one of the three landmarks in the environment.
We observe that behaviors learnt using \iiql are not only capable of addressing complex tasks (cf. Table~\ref{table. task reward results}) but also interpretable and match diverse expert intents.
This qualitative analysis also sheds light on superior intent inference performance of \iiql (cf. Table~\ref{table. inference}).
In contrast, \oiql does not effectively learn interpretable intent-driven behavior.
As we posited in Sec. \ref{sec. oiql}, our experiments suggest learning the joint $Q$-function instead of learning two factored Q-functions decreases both interpretability and overall performance.
\autoref{fig: paths} also suggests that \ogail is not able to learn intent-driven behavior that match expert intent.

\section{Conclusion}
We introduce \iiql, a novel algorithm for learning intent-driven expert behavior from heterogeneous demonstrations.
By employing an EM-style approach and implementing factored optimizations, \iiql successfully captures time-varying intents and intent-driven behavior even in complex and continuous tasks.
Our contribution includes a theoretical analysis on the convergence properties of \iiql and empirical results showcasing that \iiql outperforms the state-of-the-art baselines.

Our work also motivates several directions of future work.
The key limitation of our work is the set of intent has to be finite and known.
Moreover, when dealing with a large number of intents, the E-step can become slow.
A worthwhile future direction can be exploring methods to relax these assumptions regarding intents. 
Additionally, while intents tend to change slower compared to actions, \iiql does not explicitly leverage this characteristic.
In future work, we hope to develop methods to incorporate this prior knowledge of intent dynamics into the learning process.
Building on \iiql, we also plan to extend this work to multi-agent settings where intent of each agent can be influenced by other agents.

\begin{acks}
We thank the anonymous reviewers for their feedback.
This research was supported in part by NSF award $\#2205454$, ARO CA\# W911NF-20-2-0214, and Rice University funds.
\end{acks}
\clearpage
\balance


\clearpage
\ifarxiv
\appendix
\onecolumn

\section{Proofs}
\subsection{Proposition \ref{thm. occupancy measure}}
Let us represent the policy and the intent transition probabilities of each AMM model as $\mathcal{N} = (\pi, \zeta)$ and $\mathcal{N}' = (\pi', \zeta')$, respectively.

We will first show the \textit{if} direction: 
\begin{align*}
    \rho_{\mathcal{N}}(s, a, x, x^-) = \rho_{\mathcal{N}'}(s, a, x, x^-) \Longrightarrow \left(\rho_{\mathcal{N}}(s, a, x) = \rho_{\mathcal{N}'}(s, a, x) \text{ and } \rho_{\mathcal{N}}(s, a, x) = \rho_{\mathcal{N}'}(s, a, x)\right).
\end{align*}
Since $\rho(s, a, x) = \sum_{x^-} \rho(s, a,x, x^-)$ and $\rho_{\mathcal{N}}(s, a, x, x^-) = \rho_{\mathcal{N}'}(s, a, x, x^-)$, we can derive the following relationships:
\begin{align*}
    \rho_{\mathcal{N}}(s, a, x) = \sum_{x^-} \rho_{\mathcal{N}}(s, a, x, x^-) = \sum_{x^-} \rho_{\mathcal{N}'}(s, a, x, x^-) = \rho_{\mathcal{N}'}(s, a, x).
\end{align*}
Similarly, from $\rho(s, x, x^-) = \sum_{a} \rho(s, a, x, x^-)$, we can derive:
\begin{align*}
    \rho_{\mathcal{N}}(s, x, x^-) = \sum_{a} \rho_{\mathcal{N}}(s, a, x, x^-) = \sum_{a} \rho_{\mathcal{N}'}(s, a, x, x^-) = \rho_{\mathcal{N}'}(s, x, x^-).
\end{align*}
Thus, we get $\rho_{\mathcal{N}}(s, a, x) = \rho_{\mathcal{N}'}(s, a, x)$ and $\rho_{\mathcal{N}}(s, x, x^-) = \rho_{\mathcal{N}'}(s, x, x^-)$, and this completes the proof of the \textit{if}-direction.

Next, we will prove the \textit{if-only} direction: 
\begin{align*}
    \left(\rho_{\mathcal{N}}(s, a, x) = \rho_{\mathcal{N}'}(s, a, x) \text{ and } \rho_{\mathcal{N}}(s, x, x^-) = \rho_{\mathcal{N}'}(s, x, x^-)\right) \Longrightarrow \rho_{\mathcal{N}}(s, a, x, x^-) = \rho_{\mathcal{N}'}(s, a, x, x^-).
\end{align*}
Since $\rho(s, x) = \sum_a \rho(s, a, x)$ and $\rho_{\mathcal{N}}(s, a, x) = \rho_{\mathcal{N}'}(s, a, x)$, we can derive the following:
\begin{align}
    \rho_{\mathcal{N}}(s, x) = \sum_a \rho_{\mathcal{N}}(s, a, x) = \sum_a \rho_{\mathcal{N}'}(s, a, x)
    = \rho_{\mathcal{N}'}(s, x).  \label{eq. sx measure}
\end{align}
Then, from $\rho(s, a, x) = \pi(a|s, x)\rho(s, x)$,
\begin{align*}
    \pi(a|s, x)\rho_{\mathcal{N}}(s, x) &= \rho_\mathcal{N}(s, a, x) \\
    &= \rho_{\mathcal{N}'}(s, a, x) \hspace{15ex} (\text{by the condition})\\
    &= \pi'(a|s, x)\rho_{\mathcal{N}'}(s, x) \\
    &= \pi'(a|s, x)\rho_{\mathcal{N}}(s, x). \hspace{9ex} (\text{by Eq. \ref{eq. sx measure}})
\end{align*}
Therefore, $\pi = \pi'$.
\\
Similarly, since $\rho(s, x^-) = \sum_{x} \rho(s, x, x^-)$ and $\rho_{\mathcal{N}}(s, x, x^-) = \rho_{\mathcal{N}'}(s, x, x^-)$, we can get the following equations:
\begin{align}
    \rho_{\mathcal{N}}(s, x^-) = \sum_x \rho_{\mathcal{N}}(s, x, x^-)=\sum_x \rho_{\mathcal{N}'}(s, x, x^-)=\rho_{\mathcal{N}'}(s, x^-). \label{eq. sx- measure}
\end{align}
Also, from $\rho(s, x, x^-)=\zeta(x|s, x^-)\rho(s, x^-)$, 
\begin{align*}
    \zeta(x|s, x^-)\rho_{\mathcal{N}}(s, x^-) &= \rho_\mathcal{N}(s, x, x^-) \\
    &= \rho_{\mathcal{N}'}(s, x, x^-) \hspace{15ex} (\text{by the condition})\\
    &= \zeta'(x|s, x^-)\rho_{\mathcal{N}'}(s, x^-)\\
    &= \zeta'(x|s, x^-)\rho_{\mathcal{N}}(s, x^-). \hspace{9ex} (\text{by Eq. \ref{eq. sx- measure}})
\end{align*}
Therefore,   $\zeta = \zeta'$. 

Now since $\mathcal{N} = (\pi, \zeta) = (\pi', \zeta') = \mathcal{N}'$, by one-to-one correspondence between $\mathcal{N}$ and $\rho_\mathcal{N}(s, a, x, x^-)$, we can conclude $\rho_{\mathcal{N}}(s, a, x, x^-) = \rho_{\mathcal{N}'}(s, a, x, x^-)$. This completes the proof of Proposition \ref{thm. occupancy measure}.

\subsection{Lemma \ref{thm. pi objective}}
We first define the un-normalized $(s^-, a^-, x^-)$-occupancy measure as:
\begin{align*}
    \tilde{\rho}^-(s, a, x) \doteq \sum_{t=0}^\infty \gamma^t p(s^{t-1}=s, a^{t-1}=a, x^{t-1}=x|\mathcal{N}, \mathcal{M})
\end{align*}
where $p(s^{t-1}=s, a^{t-1}=a, x^{t-1}=x|\mathcal{N}, \mathcal{M})$ at time $t=0$ is defined as 0. Then, the following relationship holds between $\tilde{\rho}^-$ and un-normalized $(s, a, x)$-occupancy measure $\tilde{\rho}(s, a, x)$:    $$\tilde{\rho}^-(s, a, x) = \gamma \tilde{\rho}(s, a, x)$$
Thus, when normalized, we can derive: 
\begin{align}
    \rho^-(s, a, x) = \rho(s, a, x) \label{eq. prev sax occupancy}
\end{align}

Given the estimate $\rho_E^n(s, a, x)$ and the AMM model $\mathcal{N}^n$,  we estimate $(s, a, x, s^-, a^-, x^-)$-occupancy measure of the expert at the $n$-th iteration as follows:
\begin{align*}
    \check{\rho}_E^n(s, a, x, s^-, a^-, x^-) &= \pi^n(a|s, x)\zeta^n(x|s, x^-) T(s|s^-, a^-) {\rho^-_E}^n(s^-, a^-, x^-) \\
    &=\pi^n(a|s, x)\zeta^n(x|s, x^-) T(s|s^-, a^-) {\rho_E}^n(s^-, a^-, x^-). \qquad \text{(by Eq. \ref{eq. prev sax occupancy})}
\end{align*}
For brevity, we denote $\check{\rho}_E^n$ simply as $\rho_E^n$ in this section.

Then, given $\zeta^n$ is fixed, we can denote the f-divergence of $(s, a, x, s^-, a^-, x^-)$-occupancy measure as:
\begin{align*}
    L_{saxs^-a^-x^-}^n &\doteq D_f (\rho(s, a, x, s^-, a^-, x^-) || \rho_E^n(s, a, x, s^-, a^-, x^-)) \\
    &= \sum_{s, a, x, s^-, a^-, x^-} \rho_E^n (\cdot) f\left(\frac{\rho(s, a, x, s^-, a^-, x^-)}{ \rho_E^n(s, a, x, s^-, a^-, x^-)}\right)\\
    &= \sum_{s, a, x, s^-, a^-, x^-} \rho_E^n (\cdot) f\left(\frac{\pi(a|s, x)\zeta^n(x|s, x^-) T(s|s^-, a^-)\rho(s^-, a^-, x^-)}{\pi^n(a|s, x)\zeta^n(x|s, x^-) T(s|s^-, a^-) {\rho_E}^n(s^-, a^-, x^-)}\right)\\
    &= \sum_{s, a, x, s^-, a^-, x^-} \rho_E^n (\cdot) f\left(\frac{\pi(a|s, x)\rho(s^-, a^-, x^-)}{\pi^n(a|s, x)\rho_E^n(s^-, a^-, x^-)}\right).
\end{align*}
Here, we use $\sum$ for simplicity, assuming all variables are finite. To accommodate variables in continuous spaces, we can replace this with an integral ($\int$), defined with respect to an appropriate measure.

We also denote the f-divergence of $(s^-, a^-, x^-)$-occupancy measure as: $$L_{s^-a^-x^-}^n \doteq \sum_{s, a, x} {\rho^-}_E^n(s, a, x) f(\frac{\rho^-(s,a,x)}{{\rho^-}_E^n(s,a,x)}).$$
Then, we can derive the following relationship between $L_{s^-a^-x^-}^n$ and $L_{saxs^-a^-x^-}^n$:
\begin{align}
    L_{s^-a^-x^-}^n &= \sum_{s, a, x} {\rho^-}_E^n(\cdot) f(\frac{\rho^-_{\pi_\theta}(s,a,x)}{{\rho^-}_E^n(s,a,x)}) \nonumber \\
    &= \sum_{s^-, a^-, x^-} \rho_E^n(\cdot) f(\frac{\rho_{\pi_\theta}(s^-,a^-,x^-)}{\rho_E^n(s^-,a^-,x^-)}) \nonumber \\
    &= \sum_{s, a, x, s^-, a^-, x^-} \rho_E^n(\cdot) f(\frac{\pi^{n}(a|s, x) \rho_{\pi_\theta}(s^-,a^-,x^-)}{\pi^n(a|s, x) \rho_E^n(s^-,a^-,x^-)}) \nonumber \\
    &= \sum_{s, a, x, s^-, a^-, x^-} \rho_E^n(\cdot) \left\{ f(\frac{\pi^{n}(a|s, x) \rho_{\pi_\theta}(s^-,a^-,x^-)}{\pi^n(a|s, x) \rho_E^n(s^-,a^-,x^-)}) - f(\frac{\pi_\theta(a|s, x) \rho_{\pi_\theta}(s^-a^-x^-)}{\pi^n(a|s, x) \rho_E^n(s^-a^-x^-)}) + f(\frac{\pi_\theta(a|s, x) \rho_{\pi_\theta}(s^-a^-x^-)}{\pi^n(a|s, x) \rho_E^n(s^-a^-x^-)})  \right\} \nonumber \\
    &= \sum_{s, a, x, s^-, a^-, x^-} \rho_E^n(\cdot) \left\{ f(\frac{\pi^{n}(a|s, x) \rho_{\pi_\theta}(s^-,a^-,x^-)}{\pi^n(a|s, x) \rho_E^n(s^-,a^-,x^-)}) - f(\frac{\pi_\theta(a|s, x) \rho_{\pi_\theta}(s^-a^-x^-)}{\pi^n(a|s, x) \rho_E^n(s^-a^-x^-)}) \right\}  +  L_{saxs^-a^-x^-}^n \label{eq. lemma1 summation}
\end{align}
where we omit the dependency on $\zeta^n$ from $\rho_{\pi_\theta}$ for notational convenience.

Let us denote $\delta \doteq \Delta(\theta', \theta^n)$. If $|\delta|$ is sufficiently small, we can apply the first order approximation with respect to $\theta'$ at $\theta^n$. Then, we can derive the following equalities regarding the summation term of Eq. \ref{eq. lemma1 summation}:  
\begin{align*}
    & \sum_{s, a, x, s^-, a^-, x^-} \rho_E^n(\cdot) \left\{ f(\frac{\pi^{n} \rho_{\pi_\theta}(s^-,a^-,x^-)}{\pi^n \rho_E^n(s^-,a^-,x^-)}) - f(\frac{\pi_\theta \rho_{\pi_\theta}(s^-a^-x^-)}{\pi^n \rho_E^n(s^-a^-x^-)}) \right\}  \\
    &\approx  \sum_{s, a, x, s^-, a^-, x^-} \rho_E^n(\cdot) \left\{  f'(\frac{\rho_{\pi^n}(s^-,a^-,x^-)}{\rho_E^n(s^-,a^-,x^-)}) \frac{\rho_{\pi}(s^-,a^-,x^-)'|_{\pi=\pi^n}}{\rho_E^n(s^-,a^-,x^-)} \nabla_\theta \pi \delta  - f'(\frac{\rho_{\pi^n}(s^-,a^-,x^-)}{\rho_E^n(s^-,a^-,x^-)}) \frac{(\pi \rho_{\pi}(s^-,a^-,x^-))'|_{\pi=\pi^n}}{\pi^n\rho_E^n(s^-,a^-,x^-)} \nabla_\theta\pi \delta  \right\}   \\
    &=  \sum_{s, a, x, s^-, a^-, x^-} \rho_E^n(\cdot) \left\{  f'(\frac{\rho_{\pi^n}(s^-,a^-,x^-)}{\rho_E^n(s^-,a^-,x^-)}) \frac{\rho_{\pi}(s^-,a^-,x^-)'|_{\pi=\pi^n}}{\rho_E^n(s^-,a^-,x^-)} \nabla_\theta\pi \delta 
    - f'(\frac{\rho_{\pi^n}(s^-,a^-,x^-)}{\rho_E^n(s^-,a^-,x^-)})  \frac{\rho_{\pi^n}(s^-,a^-,x^-) + \pi^n \rho_{\pi}(s^-,a^-,x^-)'|_{\pi=\pi^n} }{\pi^n\rho_E^n(s^-,a^-,x^-)} \nabla_\theta\pi \delta  \right\}  \\
    &=  \sum_{s, a, x, s^-, a^-, x^-} - \rho_E^n(s, a, x, s^-, a^-, x^-) f'(\frac{\rho_{\pi^n}(s^-,a^-,x^-)}{\rho_E^n(s^-,a^-,x^-)}) \frac{\rho_{\pi^n}(s^-,a^-,x^-)}{\pi^n\rho_E^n(s^-,a^-,x^-)} \nabla_\theta\pi \delta  \\
    &= \sum_{s, a, x, s^-, a^-, x^-} - T(s|s^-,a^-) \zeta^n(x|s, x^-) \rho_{\pi^{n}}(s^-,a^-,x^-)  f'(\frac{\rho_{\pi^n}(s^-,a^-,x^-)}{\rho_E^n(s^-,a^-,x^-)}) \sum_i \frac{d\pi}{d\theta_i} \delta_i  \\
    &=-  \sum_i \frac{d}{d\theta_i} \sum_{s, a, x, s^-, a^-, x^-} \pi_\theta(a|s, x) T(s|s^-,a^-) \zeta^n(x|s, x^-) \rho_{\pi^{n}}(s^-,a^-,x^-) f'(\frac{\rho_{\pi^n}(s^-,a^-,x^-)}{\rho_E^n(s^-,a^-,x^-)}) \delta_i  \\
    &=-  \sum_i \frac{d}{d\theta_i} \sum_{s^-, a^-, x^-} \rho_{\pi^{n}}(s^-,a^-,x^-) f'(\frac{\rho_{\pi^n}(s^-,a^-,x^-)}{\rho_E^n(s^-,a^-,x^-)}) \delta_i  \\
    &= 0 \qquad
\end{align*}
where the last equality holds because the inner summation term is not a function of $\theta_i$. Therefore, under the given condition, Eq. \ref{eq. lemma1 summation} leads to the following relationship:
\begin{align}
    L_{s^-a^-x^-}^n = L_{saxs^-a^-x^-}^n  \label{eq. L relationship}
\end{align}
Due to convexity of $f$, 
\begin{align}
    L_{saxs^-a^-x^-}^n \geq L_{saxx^-}^n \geq L_{sax}^n \label{eq. D_f inequality}
\end{align}
Then, since $L_{sax}^n = L_{s^-a^-x^-}^n = L_{saxs^-a^-x^-}^n$ by Eqs. \ref{eq. prev sax occupancy} and \ref{eq. L relationship}, the leftmost term and the rightmost term of Eq. \ref{eq. D_f inequality} become equal. Thus, $L_{saxx^-}^n = L_{sax}^n$, which proves Lemma \ref{thm. pi objective}.

\subsection{Lemma \ref{thm. zeta objective}}
Given the estimate $\rho_E^n(s, x, x^-)$ and the AMM model $\mathcal{N}^n$, we define an approximate of the expert's $(s, a, x, x^-)$-occupancy measure at the $n$-th iteration as:
\begin{align*}
    \grave{\rho}_E^n(s, a, x, x^-) = \pi^n(a|s, x) \rho_E^n(s, x, x^-).
\end{align*}
Also, since $\pi^{n+1}(a|s, x)$ is given for M2-step, the $(s, a, x, x^-)$-occupancy measure of the current imitation learner $\mathcal{N}=(\pi^{n+1}, \zeta_\phi)$ can be expressed as:
\begin{align*}
    \rho_{\mathcal{N}}(s, a, x, x^{-}) = \pi^{n+1}(a|s, x) \rho_{\pi^{n+1},\zeta_\phi}(s, x, x^{-}).
\end{align*}
For notational convenience, we will omit $\pi^{n+1}$ from $\rho_{\pi^{n+1},\zeta_\phi}(s, x, x^{-})$ and express it as $\rho_{\zeta_\phi}(s, x, x^{-})$.
Then, we can derive the following relationship between $L_{sxx^-}^n$ and $L_{saxx^-}^n$:
\begin{align}
    L_{sxx^-}^n &\doteq D_f(\rho_{\zeta_\phi}(s, x, x^-) || \rho_E^n(s, x, x^-)) \nonumber \\
    &= \sum_{s, x, x^-} \rho_E^n(s, x, x^-) f\left( \frac{\rho_{\zeta_\phi}(s, x, x^-)}{\rho_E^n(s, x, x^-)} \right) \nonumber\\
    &= \sum_{s, a, x, x^-} \rho_E^n(s, a, x, x^-) f\left( \frac{\pi^n(a|s,x)\rho_{\zeta_\phi}(s, x, x^-)}{\pi^n(a|s,x)\rho_E(s, x, x^-)} \right) \nonumber\\
    &= \sum_{s, a, x, x^-} \rho_E^n (s, a, x, x^-) \left\{ f( \frac{\pi^n(a|s,x)\rho_{\zeta_\phi}(s, x, x^-)}{\pi^n(a|s,x)\rho_E^n(s, x, x^-)}) - f( \frac{\pi^{n+1}(a|s,x)\rho_{\zeta_\phi}(s, x, x^-)}{\pi^n(a|s,x)\rho_E^n(s, x, x^-)})  + f( \frac{\pi^{n+1}(a|s,x)\rho_{\zeta_\phi}(s, x, x^-)}{\pi^n(a|s,x)\rho_E^n(s, x, x^-)}) \right\} \nonumber\\
    &= \sum_{s, a, x, x^-} \rho_E^n (s, a, x, x^-) \left\{ f( \frac{\pi^n(a|s,x)\rho_{\zeta_\phi}(s, x, x^-)}{\pi^n(a|s,x)\rho_E^n(s, x, x^-)}) - f( \frac{\pi^{n+1}(a|s,x)\rho_{\zeta_\phi}(s, x, x^-)}{\pi^n(a|s,x)\rho_E^n(s, x, x^-)})\right\} + L_{saxx^-}^n \label{eq. lemma2 summation}
\end{align}
Define $\delta \doteq \Delta(\theta^n, \theta^{n+1})$. Then, since $|\delta|$ is sufficiently small by the condition, we take the first-order approximation of $\theta^{n+1}$ at $\theta^n$ in the summation term.
\begin{align*}
    &\sum_{s, a, x, x^-} \rho_E^n (s, a, x, x^-) \left\{ f( \frac{\pi^n(a|s,x)\rho_{\zeta_\phi}(s, x, x^-)}{\pi^n(a|s,x)\rho_E^n(s, x, x^-)}) - f( \frac{\pi^{n+1}(a|s,x)\rho_{\zeta_\phi}(s, x, x^-)}{\pi^n(a|s,x)\rho_E^n(s, x, x^-)})\right\} \\
    &\approx \sum_{s, a, x, x^-} \rho_E^n (s, a, x, x^-) \left\{f'( \frac{\rho_{\zeta_\phi,\pi^n}(s, x, x^-)}{\rho_E^n(s, x, x^-)} ) \frac{\rho_{\zeta_\phi}(s, x, x^-)'|_{\pi=\pi^n}}{\rho_E^n(s, x, x^-)}\nabla_\theta\pi \delta -  f'( \frac{\rho_{\zeta_\phi,\pi^n}(s, x, x^-)}{\rho_E^n(s, x, x^-)} ) \frac{\rho_{\zeta_\phi,\pi^n}(s, x, x^-) + \pi^n \rho_{\zeta_\phi}(s, x, x^-)'|_{\pi=\pi^n}}{\pi^n\rho_E^n(s, x, x^-)}\nabla_\theta\pi \delta \right\} \\
    &= \sum_{s, a, x, x^-} \pi^n \rho_E^n (s, x, x^-)  \left\{- f'( \frac{\rho_{\zeta_\phi,\pi^n}(s, x, x^-)}{\rho_E^n(s, x, x^-)} ) \frac{\rho_{\zeta_\phi,\pi^n}(s, x, x^-)}{\pi^n\rho_E^n(s, x, x^-)}\nabla_\theta\pi \delta \right\} \\
    &= - \sum_{s, a, x, x^-} \rho_{\zeta_\phi,\pi^n}(s, x, x^-) f'( \frac{\rho_{\zeta_\phi,\pi^n}(s, x, x^-)}{\rho_E^n(s, x, x^-)} ) \nabla_\theta\pi \delta \\
    &=  - \sum_{s, a, x, x^-} \rho_{\zeta_\phi,\pi^n}(s, x, x^-) f'( \frac{\rho_{\zeta_\phi,\pi^n}(s, x, x^-)}{\rho_E^n(s, x, x^-)} ) \sum_i \frac{d}{d\theta_i} \pi \delta_i \\
    &=  - \sum_i \frac{d}{d\theta_i} \sum_{s, a, x, x^-}  \pi_\theta(a|s, x) \rho_{\zeta_\phi,\pi^n}(s, x, x^-) f'( \frac{\rho_{\zeta_\phi,\pi^n}(s, x, x^-)}{\rho_E^n(s, x, x^-)} ) \delta_i \\
    &=  - \sum_i \frac{d}{d\theta_i} \sum_{s, x, x^-} \rho_{\zeta_\phi,\pi^n}(s, x, x^-) f'( \frac{\rho_{\zeta_\phi,\pi^n}(s, x, x^-)}{\rho_E^n(s, x, x^-)} ) \delta_i \\
    &=0
\end{align*}
where the last equality holds due to the independence on $\theta_i$ of the summation term. Thus, under the given condition, we can derive from Eq. \ref{eq. lemma2 summation}: $$L_{sxx^-}^n=L_{saxx^-}^n,$$
and this proves Lemma \ref{thm. zeta objective}.

\subsection{Theorem \ref{thm. convergence}}
Let us denote the objective function at the $n$-th iteration before E-step as ${L'}_{saxx^-}^n \doteq D_f (\rho_{\pi^n, \zeta^n} (s, a, x, x^-) || \rho_E^{n-1} (s, a, x, x^-))$. Note that we assumed $\rho_E^n(s,a,x,x^-) \doteq \rho_E(s, a)p(x, x^-|s, a, \mathcal{N}^n)$ for the empirical estimate of $\rho_E$ at the $n$-th E-step. Thus, $\rho_E^{n-1} (s, a, x, x^-) = \rho_E (s, a) p(x, x^-|s, a, \mathcal{N}^{n-1})$.

Then, we can derive the following relationship regarding E-step:
\begin{align*}
    {L'}_{saxx^-}^n &\doteq D_f (\rho_{\mathcal{N}^n} (s, a, x, x^-) || \rho_E^{n-1} (s, a, x, x^-))\\
        &= \sum_{s, a, x, x^-} \rho_E(s, a) p(x, x^-|s, a, \mathcal{N}^{n-1}) f\left(\frac{\rho_{\pi^n, \zeta^n}(s, a, x, x^-)}{\rho_E (s, a) p(x, x^-|s, a, \mathcal{N}^{n-1})}\right) \\
        &\geq \sum_{s, a} \rho_E(s, a) f\left(\frac{\rho_{\pi^n, \zeta^n}(s, a)}{\rho_E (s, a)}\right) \hspace{35ex} \text{($\because$ $f$ is convex)}\\
        &=\sum_{s, a, x, x^-} \rho_E(s, a) p(x, x^-|s, a, \mathcal{N}^{n}) f\left(\frac{\rho_{\pi^n, \zeta^n}(s, a, x, x^-)}{\rho_E (s, a) p(x, x^-|s, a, \mathcal{N}^{n}) }\right)  \hspace{4ex} \text{($\because$ E-step)}\\
        &=\sum_{s, a, x, x^-} \rho_E^n(s, a, x, x^-) f\left(\frac{\rho_{\pi^n, \zeta^n}(s, a, x, x^-)}{\rho_E^n(s, a, x, x^-) }\right)\\
        &=L_{saxx^-}^n 
\end{align*}
Thus, ${L'}_{saxx^-}^n \geq L_{saxx^-}^n$. 

Under the conditions of Theorem \ref{thm. convergence}, we can derive the following inequalities regarding \pistep and \zetastep with lemmas \ref{thm. pi objective} and \ref{thm. zeta objective}:
\begin{align*}
        L_{saxx^-}^n &\doteq \sum_{s, a, x, x^-} \rho_E^n(s, a, x, x^-) f\left(\frac{\rho_{\pi^n, \zeta^n}(s, a, x, x^-)}{\rho_E^n(s, a, x, x^-) }\right)\\
        &\geq \sum_{s, a, x, x^-}  \rho_E^n(s, a,x,x^-) f\left(\frac{\rho_{\pi^{n+1}, \zeta^{n}}(s, a, x, x^-)}{\rho_E^n(s, a, x, x^-)}\right)  \hspace{5ex} \text{($\because$ Lemma \ref{thm. pi objective})}\\
        &\geq \sum_{s, a, x, x^-}   \rho_E^n(s, a,x,x^-) f\left(\frac{\rho_{\pi^{n+1}, \zeta^{n+1}}(s, a, x, x^-)}{\rho_E^n(s, a, x, x^-)}\right)  \hspace{4ex} \text{($\because$ Lemma \ref{thm. zeta objective})}\\
        &= {L'}_{saxx^-}^{n+1}.
\end{align*}
Therefore, ${L'}_{saxx^-}^n \geq L_{saxx^-}^n \geq {L'}_{saxx^-}^{n+1} \geq {L}_{saxx^-}^{n+1}$, and this proves Theorem \ref{thm. convergence}.

\section{Implementation Considerations}
\label{sec. appendix implementation considerations}
In theory, Algorithm~\ref{alg. learner} is agnostic to the parameterization used for $\pi$ and $\zeta$.
However, in practical implementations, function approximation is more desirable than exact tabular representation to tackle high-dimensional or continuous state representations.
Following \iql, in our implementation, we opt for neural networks due to their expressiveness. 
As mentioned in Sec. \ref{sec. implementation considerations}, we choose small learning rates so that the neural network parameters can be updated only by small increments and repeat \pistep and \zetastep (lines 7-8) multiple times before the subsequent E-step (line 4). 
While mathematically the estimate of $\rho_E(s, a, x, x^-)$ will immediately change whenever $\pi_\theta$ and $\zeta_\phi$ change, empirically we observe that the results of E-step are less sensitive to the incremental updates of $\theta$ and $\phi$.
We utilize the discrete version of \iql for learning $\zeta$ and $\pi$ in domains with discrete action spaces, and use its continuous version to learn $\pi$ in domains with continuous action space.
This implementation especially improves the performance in discrete domains, as \iql derives $\pi$ directly from the learnt Q-network.
As mentioned in Sec. \ref{sec. oiql}, the na\"ive application of \iql to Eq. \ref{eq. main objective} requires additional soft-actor-critic (SAC) optimization \cite{haarnoja2018soft}, which often requires reparameterization tricks for practical implementation. 
While our implementation employs \iql for both the M-steps, \iiql itself does not limit the choice of imitation learning algorithms for Eq. \ref{eq. m-step pi objective} and \ref{eq. m-step zeta objective}.

\newcommand{\myurl}[1]{{\color{blue}\url{#1}}}

\section{Implementation Details}
We use Python 3.8 and PyTorch 2.0.0 to develop all neural network-based algorithms.

\subsection{Experimental Domains}
In Fig. \ref{fig: another mutligoals},we show the screenshots of \twodim domains for $n=2, 4, 5$.

\subsection{Baseline Implementation}
We utilize the implementations provided by \myurl{https://github.com/id9502/Option-GAIL} for \ogail and \myurl{https://github.com/Div99/IQ-Learn} for \iql. For Behavior Cloning, we use the implementation available at \myurl{https://github.com/HumanCompatibleAI/imitation}. As the original implementations of \ogail and \iql are not compatible with discrete state spaces, we have made modifications, which convert a discrete state into its one-hot encoding, to enable them to work with discrete-state-space environments. In addition, since \ogail implementation does not offer a discrete actor, we have additionally implemented an actor that outputs Categorical distribution to test \ogail with discrete environments.

\subsection{\oiql Implementation}
In \oiql, a neural network is employed for the Q-function, which takes $s, a, x$ and $x^-$ as its input. When dealing with intents from the dataset $\tilde{D}$, since they are discrete, we one-hot encoded them before feeding them into the neural networks. Discrete states and actions are also one-hot encoded. Similarly, the policy network and the intent-transition network are constructed, taking $(s, x)$ and $(s, x^-)$ as their inputs, respectively, with one-hot encoding applied if necessary.

\begin{figure*}[t]
  \centering
  \begin{subfigure}[t]{0.3\linewidth}
      \centering
      \includegraphics[width=0.7\textwidth, frame]{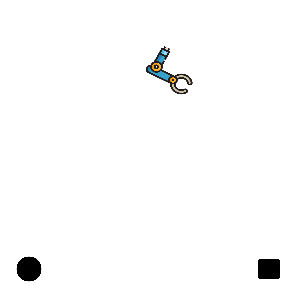}
      \caption{\twodim-$2$}
      \label{fig. mg2}
  \end{subfigure}
  \hfill
  \begin{subfigure}[t]{0.3\linewidth}
      \centering
      \includegraphics[width=0.7\textwidth, frame]{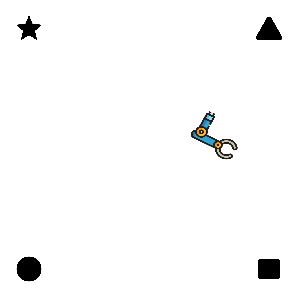}
      \caption{\twodim-$4$}
      \label{fig: mg4}
  \end{subfigure}
  \hfill
    \begin{subfigure}[t]{0.3\linewidth}
      \centering
      \includegraphics[width=0.7\textwidth, frame]{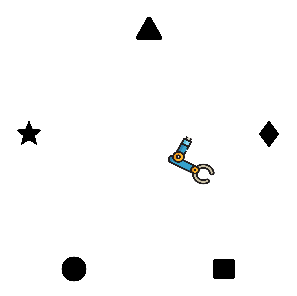}
      \caption{\twodim-$5$}
      \label{fig. mg5}
  \end{subfigure}
  \captionsetup{subrefformat=parens}
  \caption{\twodim domains that include different number of intents.}
  \label{fig: another mutligoals}
\end{figure*}

During the optimization of the SAC objective, it is essential to sample $a$ and $x$ from the policy network and the intent-transition network to compute the gradient of the $f$-divergence in Eq. \ref{eq. joint sac}. To achieve this, we employ the Gumbel-softmax reparameterization trick to maintain the gradients of $x$ generated from the intent-transition policy: $x \sim g_\phi (\mu; s, x^-)$, where $\mu \sim \textsc{Uniform}$. As for actor outputs, we utilize either the Gaussian reparameterization or the Gumbel-softmax reparameterization, depending on whether the action space is continuous or discrete: $x \sim f_\theta (\epsilon; s, x)$, where $\epsilon \sim \textsc{Normal}~\text{or}~\textsc{Uniform}$. Thus, we optimize Eq. \ref{eq. joint sac} by representing it as follows (e.g., for continuous action space):
\begin{align*}
    J(\theta, \phi) = E_{(s, x^-) \sim \tilde{D}, \epsilon \sim \textit{N}, \mu \sim \textit{U}}\left[ \log \pi_\theta \left(f_\theta (\epsilon)\middle| s, g_\phi (\mu)\right) + \log \zeta_\phi \left( g_\phi (\mu)\middle|x-, s\right)   - Q\left(f_\theta (\epsilon), g_\phi (\mu), x^-, s)\right)\right]
\end{align*}
where $\textit{N}$ denotes \textsc{Normal}, $\textit{U}$ denotes \textsc{Uniform}, $g_\phi (\mu) \doteq g_\phi(\epsilon|s, x^-)$, and  $f_\theta (\epsilon)\doteq f_\theta (\epsilon|s, g_\phi (\mu))$.

\subsection{\iiql Implementation}
Similar to \ogail implementation, \iiql employs separate Q-functions for each intent. In other words, we structure the Q-function of the intent-informed MDP as $Q(s, x, a) = Q_x(s, a)$ where $Q_x$ represents a neural network dedicated to each intent. Similarly, we create subdivided Q-functions for the intent-transition MDP as $Q(s, x, x^-) = Q_{x^-}(s, x)$, where $Q_{x^-}(s, x)$ is also a separate neural network. Furthermore, we design dedicated neural networks $\pi_x(a|s)$ for each intent, to represent the policy function $\pi(a|s, x)$.

\subsection{Hyperparameters}
For both \oiql, and \iiql, we maintain a replay buffer size of 50,000 for all experiments. Network updates occur once every 5 exploration steps and a batch size of 256 is utilized. An E-step is performed once every 200 updates of $\pi$ and $\zeta$. For Q-function corresponding to $\zeta$, we employ the ``value'' method as \iql's 2nd term loss across all domains. As for the learning rate of Q-functions, we set $3\cdot 10^{-4}$ for all configurations and domains.
For other domain-specific hyperparameters, please refer to Table \ref{table. hyperparameters}.

\newcommand{\Temp}{\textit{Temp}}
\newcommand{\Qlr}{\textit{Q-LR}}
\newcommand{\zetaTemp}{\textit{$\zeta$-Temp}}
\newcommand{\zetaQlr}{\textit{$\zeta$-Q-LR}}
\newcommand{\zetaAlr}{\textit{$\zeta$-A-LR}}
\newcommand{\piTemp}{\textit{$\pi$-Temp}}
\newcommand{\piQlr}{\textit{$\pi$-Q-LR}}
\newcommand{\piAlr}{\textit{$\pi$-A-LR}}
\newcommand{\iqloss}{\textit{IQL-2nd}}
\newcommand{\iqdiv}{\textit{IQ-div}}
\newcommand{\piiqloss}{\textit{$\pi$-IQL-2nd}}
\newcommand{\piiqdiv}{\textit{$\pi$-$D_f$}}
\newcommand{\zetaiqloss}{\textit{$\zeta$-IQL-2nd}}
\newcommand{\zetaiqdiv}{\textit{$\zeta$-$D_f$}}

\begin{table}[ht]
\centering
\caption{Hyperparameters for \oiql and \iiql. }
\label{table. hyperparameters}
\begin{tabular}{lccccccccc} \toprule
                 & \multicolumn{4}{c}{\bf \oiql}  & \multicolumn{5}{c}{\bf \iiql} \\
                 \cmidrule(lr){2-5} \cmidrule(lr){6-10}
\textbf{Domain}     & \Temp  & \zetaAlr &  \piAlr & \iqloss & \zetaTemp & \zetaiqdiv  & \piTemp & \piAlr & \piiqloss \\ 
\midrule
\twodimshort-n suite & 0.01  & 1e-4 & 1e-4 & value & 0.01  & $\chi^2$ & 0.01  & 1e-4  & value  \\
\single              & 0.01  & 1e-4 & 1e-4 & value & 0.01  & $\chi^2$ & 0.01  & 1e-4  & value \\
\movers              & 0.01  & 1e-4 & 1e-4 & value & 0.01  & $\chi^2$ & 0.01  & 1e-4  & value \\
Hopper               & 0.01  & 3e-5 & 3e-5 & v0    & 0.001 & $\chi^2$ & 0.01  & 3e-5  & v0 \\
HalfCheetah          & 0.01  & 3e-5 & 3e-5 & value & 0.01  & TV       & 0.01  & 3e-5  & value  \\
Walker               & 0.01  & 3e-5 & 3e-5 & v0    & 0.001 & $\chi^2$ & 0.01  & 3e-5  & v0  \\
Ant                  & 0.001 & 3e-5 & 3e-5 & value & 0.01  & TV       & 0.001 & 3e-5  & value  \\
Humanoid             & 1     & 3e-5 & 3e-5 & v0    & 0.001 & $\chi^2$ & 1     & 3e-5  & v0  \\
AntPush              & 0.001 & 3e-5 & 3e-5 & value & 0.01  & TV       & 0.001 & 3e-5  & value  \\
\bottomrule  
\multicolumn{5}{l}{
\makecell[l]{* \Temp: temperature parameter for \oiql \\
* \zetaTemp: temperature parameter for $\zeta$  \\
* \piTemp: temperature parameter for $\pi$ \\
* \zetaAlr: learning rate for $\zeta$-network}} &
\multicolumn{5}{l}{
\makecell[l]{* \piAlr: learning rate for $\pi$-network \\
* \zetaiqdiv: the function of $D_f$ \\
* \iqloss: \iql 2nd-term  loss \\
* \piiqloss: \iql 2nd-term loss for $\pi$.}}
\end{tabular}
\end{table}

\section{Additional Results}
In Fig. \ref{fig: learning curve}, we plot the task rewards achieved according to the number of exploration steps taken during training for the domains \twodim-$5$, \texttt{Hopper-v2}, and \texttt{Humanoid-v2}. Note that while \iql and \oiql exhibit faster convergence and perform well with intent-agnostic expert demonstrations, they underperform \iiql with intent-driven demonstrations. When compared to \ogail, \iiql shows faster convergence and achieves higher rewards in both intent-agnostic and intent-driven demonstrations.

\begin{figure*}[t]
  \centering
  \begin{subfigure}[t]{0.3\linewidth}
      \centering
      \includegraphics[width=\textwidth]{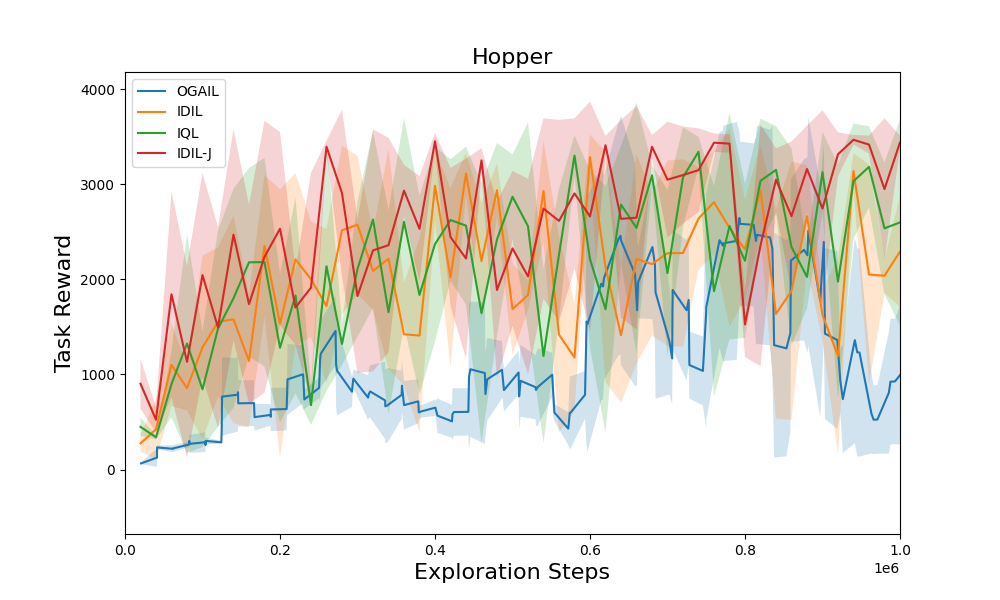}
      \caption{\texttt{Hopper}}
      \label{fig. hopper curve}
  \end{subfigure}
  \hfill
  \begin{subfigure}[t]{0.3\linewidth}
      \centering
      \includegraphics[width=\textwidth]{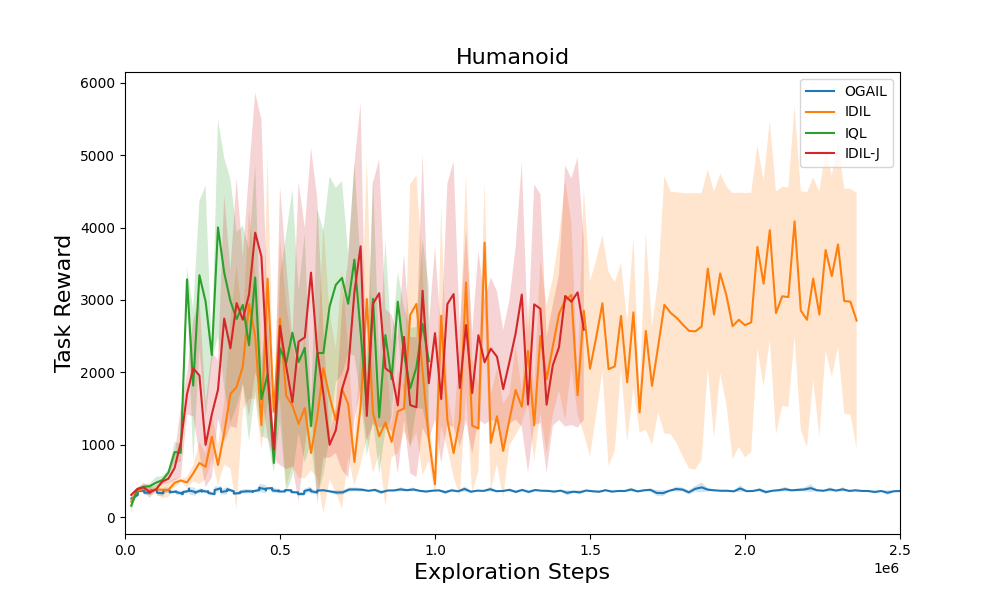}
      \caption{\texttt{Humanoid}}
      \label{fig: humanoid curve}
  \end{subfigure}
  \hfill
    \begin{subfigure}[t]{0.3\linewidth}
      \centering
      \includegraphics[width=\textwidth]{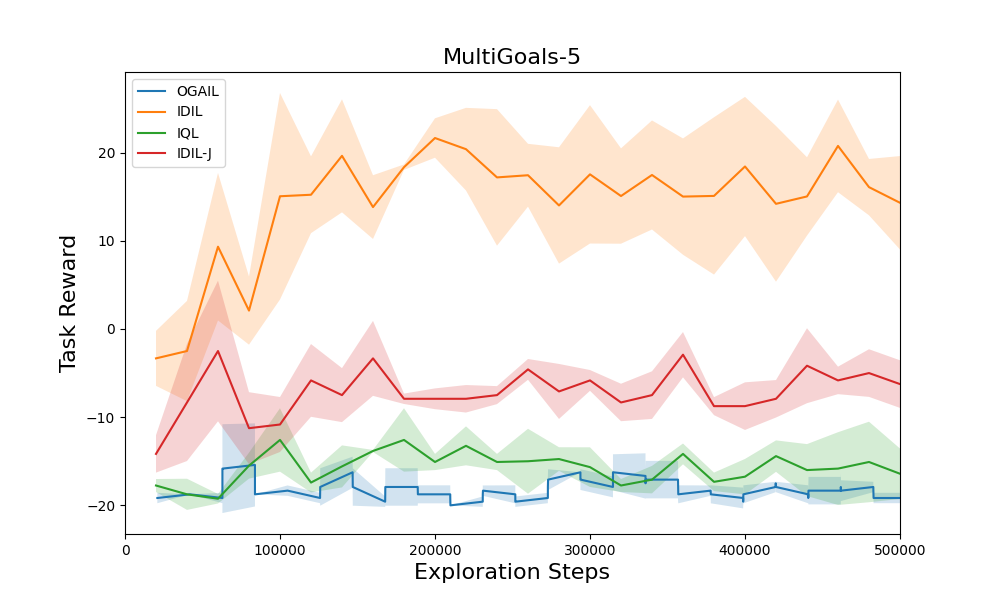}
      \caption{\twodim-$5$}
      \label{fig. mg5 curve}
  \end{subfigure}
  \captionsetup{subrefformat=parens}
  \caption{The average task returns vs. the number of exploration steps. For \texttt{Humanoid}, training was stopped once each algorithm had converged, having reached expert rewards.}
  \label{fig: learning curve}
\end{figure*}

\section{Concepts Related to Intent}
A large body of work aims to learn intrinsic rewards from demonstrations to elucidate the rationale behind an expert's behavior \cite{abbeel2004apprenticeship, ziebart2008maximum, arora2021survey}. Since our work also considers intent to explain expert behavior, one might associate intent with intrinsic rewards. However, the notion of intent in our work is distinct from that of intrinsic rewards. In our setting, experts may exhibit markedly diverse behaviors even within the same task context. While intrinsic rewards are effective at explaining one representative behavior among them, they are insufficient for expressing a range of diverse behaviors. On the other hand, intent allows us to categorize different expert behaviors into several groups, thereby helping us capture a broader spectrum of expert behavior from demonstrations.

Some works integrate other concepts of latent states, such as subtasks, subgoals, strategies, or mental states, into agent models to capture diverse expert behavior \cite{jing2021adversarial, sharma2018directed, unhelkar2019learning, li2017infogail}. Despite the varied terminology, these notions practically align with the intent in our work in that when an expert exhibits different behaviors in the same context, they are linked to distinct categories. While many studies consider only a finite set of latent states, \citet{lee2020learning} and \citet{orlov2022factorial} explore continuous and multidimensional latent states, respectively. Although our work does not specifically address these variants, they could be interpreted similarly to intents. Extending our intent representation to include these variants will be an interesting future study.

\fi

\end{document}